\documentclass[conference,a4paper]{IEEEtran}
\usepackage[utf8]{inputenc}
\usepackage[T1]{fontenc}
\usepackage{url}			 
\usepackage{cite}            

\usepackage{hyperref}
\hypersetup{ 
    colorlinks=true, 
    linkcolor=black, 
    citecolor=black, 
    urlcolor=black, 
    pdftitle={Your Title}, 
    pdfauthor={Your Name}, 
    bookmarksnumbered=true,
    bookmarksopen=true,
    breaklinks=true,
}
\usepackage[cmex10]{amsmath}  

\usepackage{mleftright}       

\usepackage{cgitIEEE}  
\mleftright

\usepackage{graphicx}         
\usepackage{booktabs}         

\usepackage{algorithmicx}     

\usepackage[caption=false,font=footnotesize]{subfig}

\usepackage[dvipsnames]{xcolor}

\IEEEoverridecommandlockouts

\begin{document}

\title{Generalization Error of $f$-Divergence Stabilized Algorithms via Duality
\thanks{This work is supported by the University of Sheffield ACSE PGR scholarships, the Inria Exploratory Action -- This work is supported in part by the European Commission through the H2020-MSCA-RISE-2019 project 872172; the French National Agency for Research (ANR)  through the Project ANR-21-CE25-0013 and the project ANR-22-PEFT-0010 of the France 2030 program PEPR R\'eseaux du Futur.}
}




\author{%
 \IEEEauthorblockN{Francisco Daunas\IEEEauthorrefmark{1}\IEEEauthorrefmark{2},
                    I{\~n}aki Esnaola \IEEEauthorrefmark{1}\IEEEauthorrefmark{3},
                    Samir M. Perlaza\IEEEauthorrefmark{2}\IEEEauthorrefmark{3}\IEEEauthorrefmark{5},
                    and Gholamali Aminian\IEEEauthorrefmark{4}}
   \IEEEauthorblockA{\IEEEauthorrefmark{1}%
                     Sch. of Electrical and Electronic Engineering, University of Sheffield,
                     Sheffield, U.K.
                     \{jdaunastorres1, esnaola\}@sheffield.ac.uk}
   \IEEEauthorblockA{\IEEEauthorrefmark{2}%
                     INRIA,
                     Centre Inria d'Universit\'e C\^ote d'Azur,
                     Sophia Antipolis, France.
                     samir.perlaza@inria.fr}
   \IEEEauthorblockA{\IEEEauthorrefmark{3}%
                     ECE Dept., Princeton University, Princeton,
                     08544 NJ, USA.}
   \IEEEauthorblockA{\IEEEauthorrefmark{4}%
                     Alan Turing Institute, London, United Kingdom. gaminian@turing.ac.uk}
   \IEEEauthorblockA{\IEEEauthorrefmark{5}%
                     GAATI, Universit\'e de la Polyn\'esie Fran\c{c}aise,
                     Faaa, French Polynesia.}
}
%

\maketitle
\begin{abstract}
  THIS PAPER IS ELIGIBLE FOR THE STUDENT PAPER AWARD.
    The solution to empirical risk minimization with $f$-divergence regularization (ERM-$f$DR) is extended to constrained optimization problems, establishing conditions for equivalence between the solution and constraints. A dual formulation of ERM-$f$DR is introduced, providing a computationally efficient method to derive the normalization function of the ERM-$f$DR solution. This dual approach leverages the Legendre-Fenchel transform and the implicit function theorem, enabling explicit characterizations of the generalization error for general algorithms under mild conditions, and another for ERM-$f$DR solutions.
\end{abstract}

%
%
\section{Introduction}
\label{sec:introduction}
In statistical learning, the classical empirical risk minimization (ERM) problem~\cite{vapnik1964perceptron,vapnik1992principles} is transformed by minimizing the expected empirical risk over a subset of all probability measures defined on the set of models. A regularization term is added to this expected empirical risk, often expressed as a  \textit{statistical distance} between the optimization measure and a reference measure~\cite{raginsky2016information,russo2019much}.
The reference measure is not necessarily a probability measure, and can be a $\sigma$-finite measure, as shown in~\cite{perlaza2024ERMRER} and~\cite{Perlaza_ISIT_2022}. This paper focuses on a family of statistical distances known as $f$-divergences, introduced in~\cite{renyi1961measures} and further developed in~\cite{sason2016fdivergence} and~\cite{csiszar1967information}.

ERM problems with $f$-divergence regularization (ERM-$f$DR) have been previously explored in~\cite{teboulle1992entropic} and~\cite{beck2003mirror} for discrete cases, and in a more general setting in~\cite{alquier2021non} as a non-exponentially weighted aggregation method. Recent work, such as~\cite{Perlaza-ISIT2024a}, expanded the set of $f$-divergences for which explicit solutions can be derived.
The particular cases in which $f(x) = x\log(x)$ and $f(x) = - \log(x)$ are known as \mbox{Type-I} and \mbox{Type-II} ERM with relative entropy regularization (ERM-RER) are thoroughly studied in~\cite{daunas2024TITAsymmetry, Perlaza-ISIT2023a,perlaza2024Generalization}. Relative entropy regularization has also been studied in the context of the worst-case data-generating probability measure in~\cite{zou2024WorstCase} and~\cite{zou2024generalization}. 
However, a key limitation of these results is the difficulty in evaluating the normalization function of the solution. 
The computation of the normalization function involves determining the Lagrange multipliers, which depend on the regularization factor in the ERM-$f$DR problem formulation. These multipliers are represented as the normalization function in~\cite{Perlaza-ISIT2024a}, a concept akin to the partition function in statistics. 
Computing this normalization function requires evaluating the empirical risk over all possible models in the support of the reference measure, which is considered $\#P$-hard~\cite{bulatov2005complexity}. For instance, in the case of the widely used relative entropy~\cite{kullback1951information}, its asymmetry yields two distinct formulations: \mbox{Type-I} and \mbox{Type-II} ERM with relative entropy regularization (ERM-RER)~\cite{daunas2024TITAsymmetry, Perlaza-ISIT2023a,perlaza2024Generalization}.

Dual problem formulations play a central role in optimization theory~\cite{rockafellar1970conjugate} and \cite{boyd2004convex}, that can offer both theoretical insights and computational advantages for optimization problems. Transforming the original (primal) problem into its dual counterpart opens the door to making use of properties such as convexity and separability, which are often not readily apparent in the primal formulation. In this context, the ERM-$f$DR problem benefits from dual formulations due to its strict convexity. By leveraging the techniques presented in~\cite{Perlaza-ISIT2024a} -- which goes along the lines of the methods in~\cite{perlaza2024ERMRER, zou2024WorstCase, InriaRR9474}; and relying on the G{\^a}teaux derivative~\cite{gateaux1913fonctionnelles} and vector space methods~\cite{luenberger1997bookOptimization}, this paper introduces a dual formulation for the ERM-$f$DR problem. The proposed dual formulation provides a convenient way to compute the normalization function, while offering operational insights into the characterization of the generalization error for statitstical learning algorithms.

This paper makes the following contributions: Section~\ref{sec:ERMfDR} extends the existing ERM-$f$DR solution to constrained optimization problems by establishing the conditions under which it also serves as the solution to the constrained problem. A dual optimization problem is introduced in Section~\ref{sec:dual} and its solution is derived, jointly with conditions provided to ensure equivalence between the dual and primal solutions.
The dual solution is used to characterize the normalization function via the implicit function theorem, connecting the ERM-$f$DR solution to the dual through the Legendre-Fenchel transform.
Finally, the connection between the Legendre-Fenchel transform and $f$-divergence regularization is used to explicitly characterize the generalization error for the \mbox{ERM-$f$DR} problem in Section~\ref{sec:GenErr}.

%
%
\section{Preliminaries}

Let $\Omega$ be an arbitrary subset of $\reals^{d}$, with $d \in \ints$, and let $\BorSigma{\Omega}$ denote the Borel $\sigma$-field on $\Omega$. The set of probability measures that can be defined upon the measurable space $\left(\Omega, \BorSigma{\Omega} \right)$ is denoted by~$\bigtriangleup(\Omega)$.
Given a probability measure $Q \in \bigtriangleup(\Omega)$ the set exclusively containing the probability measures in $\bigtriangleup(\Omega)$ that are absolutely continuous with respect to $Q$ is denoted by $\bigtriangleup_{Q}(\Omega)$. That is,
\begin{IEEEeqnarray}{rCl}
\label{DefSetTriangUp}
\bigtriangleup_{Q}(\Omega) & \triangleq & \{P\in \bigtriangleup(\Omega): P \ll Q \},
\end{IEEEeqnarray}
where the notation $P \ll Q$ stands for the measure $P$ being absolutely continuous with respect to the measure $Q$.
The Radon-Nikodym derivative of the measure $P$ with respect to $Q$ is denoted by $\frac{\diff P}{\diff Q}:\Omega\rightarrow [0,\infty)$.

Using this notation, an $f$-divergence is defined as follows.
\begin{definition}[$f$-divergence~\cite{csiszar1967information}]
\label{Def_fDivergence}
Let $f:[0,\infty)\rightarrow \reals$ be a convex function with $f(1)=0$ and $f(0) \triangleq \lim_{x\rightarrow 0^+}f(x)$.
Let $P$ and $Q$ be two probability measures on the same measurable space, with $P$ absolutely continuous with $Q$.
The $f$-divergence of $P$ with respect to $Q$, denoted by $\KLf{P}{Q}$, is
\begin{equation}
\label{EqD_f}
\KLf{P}{Q} \triangleq \int f(\frac{\diff P}{\diff Q}(\omega)) \diff Q(\omega).
\end{equation}
\end{definition}

In the case in which the function $f$ in~\eqref{EqD_f} is continuous and differentiable, the derivative of the function~$f$ is denoted by
\begin{equation}
\label{EqDefDiffF}
\dot{f}: (0, +\infty) \to \reals. 
\end{equation}
If the inverse of the function~$\dot{f}$ exists, it is denoted by 
\begin{equation}
\label{EqDefInvDiffF}
\dot{f}^{-1}: \reals \to (0, +\infty). 
\end{equation}

%
%
\section{Problem Formulation}
\label{sec:ERMfDR}
%
Let~$\set{M}$,~$\set{X}$ and~$\set{Y}$, with~$\set{M} \subseteq \reals^{d}$ and~$d \in \ints$, be sets of \emph{models}, \emph{patterns}, and \emph{labels}, respectively.
A pair $(x,y) \in \mathcal{X} \times \mathcal{Y}$ is referred to as a \emph{labeled pattern} or \emph{data point}, and a \emph{dataset} is represented by the tuple $((x_1, y_1), (x_2, y_2), \ldots,(x_n, y_n))\in ( \set{X} \times \set{Y} )^n$.
Let the function~$h: \set{M} \times \mathcal{X} \rightarrow \mathcal{Y}$ be such that the label assigned to a pattern $x \in \set{X}$ according to the model $\thetav \in \set{M}$ is $h(\thetav,x)$.
Then, given a dataset
\begin{equation}
\label{EqTheDataSet}
\vect{z} = \big((x_1, y_1), (x_2, y_2 ), \ldots, (x_n, y_n )\big)  \in ( \set{X} \times \set{Y} )^n,
\end{equation}
the objective is to obtain a model $\thetav \in \set{M}$, such that, for all $i \inCountK{n}$, the label assigned to the pattern $x_i$, which is $h(\thetav,x_i)$, is ``close'' to the label $y_i$.
This notion of ``closeness'' is formalized by the function
\begin{equation}
\label{EqRiskFunDef}
    \ell: \set{Y} \times \set{Y} \rightarrow [0, +\infty),
\end{equation}
such that the loss or risk induced by choosing the model $\thetav \in \set{M}$  with respect to the labeled pattern $(x_i, y_i)$, with $i\inCountK{n}$, is $\ell(h(\thetav,x_i),y_i)$.
The risk function $\ell$ is assumed to be nonnegative and to satisfy $\ell( y, y ) = 0$, for all $y\in\set{Y}$.

The \emph{empirical risk} induced by a model $\vect{\theta}$ with respect to the dataset $\vect{z}$ in~\eqref{EqTheDataSet} is determined by the function $\mathsf{L}_{\vect{z}}\!:\! \set{M} \rightarrow [0, +\infty)$, which satisfies
\begin{IEEEeqnarray}{rcl}
\label{EqLxy}
\mathsf{L}_{\vect{z}} (\vect{\theta} )  &\ = \ &
\frac{1}{n}\sum_{i=1}^{n}  \ell ( h(\vect{\theta}, x_i), y_i ).
\end{IEEEeqnarray}
%
%
The expectation of the empirical risk $\mathsf{L}_{\vect{z}} (\vect{\theta} )$ in~\eqref{EqLxy}, when~$\vect{\theta}$ is sampled from a probability measure $P \in \bigtriangleup(\set{M})$, is determined by the functional $\mathsf{R}_{\dset{z}}: \bigtriangleup(\set{M}) \rightarrow  [0, +\infty)$, such that
\begin{equation}
\label{EqRxy}
\foo{R}_{\dset{z}}( P ) = \int \foo{L}_{ \dset{z} } ( \thetav )  \diff P(\thetav).
\end{equation}
%

The ERM-$f$DR problem is parametrized by a probability measure $Q \in \bigtriangleup(\set{M})$, a positive real $\lambda$, and a function $f:[0,\infty)\to\reals$ that satisfies the conditions in Definition~\ref{Def_fDivergence}.
The measure $Q$ is referred to as the \emph{reference measure} and $\lambda$ as the \emph{regularization factor}.

Given the dataset~$\dset{z} \in (\set{X} \times \set{Y})^n$ in~\eqref{EqTheDataSet}, the \mbox{ERM-$f$DR} problem, with parameters~$Q$,~$\lambda$ and $f$, consists of the following optimization problem:
\begin{IEEEeqnarray}{rcl}
\label{EqOp_f_ERMRERNormal}
\min_{P \in \bigtriangleup_{Q}(\set{M})} & \quad \foo{R}_{\dset{z}} ( P ) + \lambda \KLf{P}{Q},
\end{IEEEeqnarray}
where the functional $\foo{R}_{\dset{z}}$ is defined in~\eqref{EqRxy}. The optimization problem in~\eqref{EqOp_f_ERMRERNormal} is closely related to the following optimization problem:
\vspace{-2mm}
\begin{subequations}
\label{EqOp_f_ERM_RND2}
\begin{IEEEeqnarray}{cCl}
	\min_{P \in \bigtriangleup_{Q}(\set{M})}
	& \quad & \foo{R}_{\dset{z}} ( P )\\
	\text{s.t.} 
 	& &  \Divf{P}{Q} \leq \eta,\label{EqOp_f_ERM_RND2_c_s1}
\end{IEEEeqnarray}
\end{subequations}
with $\eta \in [0,\infty)$.
The optimization problems in~\eqref{EqOp_f_ERMRERNormal} and~\eqref{EqOp_f_ERM_RND2} do not share the same solutions when for all $\thetav \in \supp Q$,  $\foo{L}_{\dset{z}}(\thetav) = c$, for some $c > 0$. More specifically, the set of solutions to the problem in~\eqref{EqOp_f_ERM_RND2} is  $\{P \in \bigtriangleup_{Q}(\set{M}): \Divf{P}{Q} \leq \eta\}$, while the set of solutions to~\eqref{EqOp_f_ERMRERNormal} is the singleton $\{Q\}$.
This distinction is mathematically significant but can be ignored in practice, as it arises only when $\foo{R}_{\dset{z}}(P)$ in~\eqref{EqRxy} is constant for all measures $P$. In order to avoid the above case, the notion of separable empirical risk functions~\cite[Definition 5]{perlaza2024ERMRER} is adopted.

\begin{definition}[Separable Empirical Risk Function \cite{perlaza2024ERMRER}]
\label{Def_SeparableLxy}
The empirical risk function $\foo{L}_{\dset{z}}$ in~\eqref{EqLxy} is said to be separable with respect to a $\sigma$-finite measure $P\in \bigtriangleup(\set{M})$, if there exists a positive real $c>0$ and two subsets $\set{M}_{1} $ and $\set{M}_{2}$ of $\set{M}$ that are nonnegligible with respect to $P$, such that for all $(\thetav_1, \thetav_2) \in \set{M}_{1} \times \set{M}_{2}$,
\vspace{-2mm}
\begin{IEEEeqnarray}{rCCCCCl}
\foo{L}_{\dset{z}}(\thetav_1)& < & c & < &  \foo{L}_{\dset{z}}(\thetav_2)	& < & \infty.
\end{IEEEeqnarray}
\end{definition}

In a nutshell, a nonseparable empirical risk function with respect to the measure $Q$ in~\eqref{EqOp_f_ERMRERNormal} satisfies
\begin{IEEEeqnarray}{rCl}
Q(\{\thetav \in \set{M}:\foo{L}_{\dset{z}}(\thetav)=a\})& = & 1,
\end{IEEEeqnarray}
for some $a > 0$.

The solutions to the ERM-$f$DR problems in~\eqref{EqOp_f_ERMRERNormal} and~\eqref{EqOp_f_ERM_RND2} are presented under the following assumptions: 
\begin{itemize}
\item[\namedlabel{assum:a}{$(a)$}] The function $f$ is strictly convex and differentiable;
\item[\namedlabel{assum:b}{$(b)$}] There exists a $\beta$ such that
\begin{subequations}
\label{EqfKrescConstrainAll}
\begin{equation}
\label{EqDefSetB}
\beta \in \left\lbrace t\in \reals: \forall \vect{\theta} \in \supp Q , 0 < \dot{f}^{-1} \left( -\frac{t + \foo{L}_{\vect{z}}(\thetav)}{\lambda} \right) \right\rbrace,
\end{equation}
and
\begin{IEEEeqnarray}{rCl}
\label{EqEqualToABigOne}
\int \dot{f}^{-1}(-\frac{\beta + \foo{L}_{\dset{z}}(\thetav)}{\lambda}) \diff Q(\thetav) & = & 1,
\IEEEeqnarraynumspace
\end{IEEEeqnarray}
where the function $\foo{L}_{\dset{z}}$ is defined in~\eqref{EqLxy}; and 
\item[\namedlabel{assum:c}{$(c)$}] The function $\foo{L}_{\dset{z}}$ in~\eqref{EqLxy} is separable with respect to the probability measure $Q$. 
\end{subequations}
\end{itemize}

Under Assumptions \ref{assum:a} and \ref{assum:b}, the solution to the optimization problem in~\eqref{EqOp_f_ERMRERNormal} was first presented in \cite[Theorem 1]{Perlaza-ISIT2024a}.
Using Assumption~\ref{assum:c}, the following theorem shows that the problems in~\eqref{EqOp_f_ERMRERNormal} and~\eqref{EqOp_f_ERM_RND2} share the same unique solution.

\begin{theorem}
\label{Theo_f_ERMRadNik}
Under Assumptions \ref{assum:a} and \ref{assum:b}, the solution to the optimization problem in~\eqref{EqOp_f_ERMRERNormal}, denoted by $\Pgibbs{P}{Q} \in \bigtriangleup_{Q}(\set{M})$, is unique, and for all $\thetav \in \supp Q$, 
\begin{equation}
\label{EqGenpdffDv}
\frac{\diff \Pgibbs{P}{Q}}{\diff Q} ( \thetav ) = \dot{f}^{-1}(-\frac{\beta + \foo{L}_{\dset{z}}(\thetav)}{\lambda}).
\end{equation}
Moreover, under Assumptions \ref{assum:a}, \ref{assum:b}, and \ref{assum:c}, if $\lambda$ in~\eqref{EqOp_f_ERMRERNormal} and $\eta$ in~\eqref{EqOp_f_ERM_RND2} satisfy
\begin{equation}
\label{EqGenDivf_eta}
\Divf{\Pgibbs{P}{Q}}{Q} = \eta,
\end{equation}
then, the probability measure $\Pgibbs{P}{Q}$ in~\eqref{EqGenpdffDv} is also the unique solution to the optimization problem in~\eqref{EqOp_f_ERM_RND2}.
\end{theorem}
\begin{IEEEproof}
	The proof is presented in Appendix~\ref{app_theo_f_ERMRadNik2}.
\end{IEEEproof}
Interestingly, the proof presented in \cite[Appendix B]{daunas2024Arxiv} for~\eqref{EqGenpdffDv}, is different from the one in \cite[Theorem 1]{Perlaza-ISIT2024a}. 
Also, note that the equality in~\eqref{EqGenpdffDv} can be written in terms of the \emph{normalization function}, introduced in~\cite{Perlaza-ISIT2024a} and defined hereunder.
\begin{definition}[Normalization Function]
The normalization function of the problem in~\eqref{EqOp_f_ERMRERNormal}, denoted by
\begin{subequations}
\label{EqDefNormFunction}
\begin{equation}
\label{EqDefMapNormFunction}
N_{Q, \dset{z}}: \set{A}_{Q,\dset{z}} \rightarrow \set{B}_{Q,\dset{z}},
\end{equation}
with $\set{A}_{Q, \dset{z}}\subseteq (0,\infty)$ and $\set{B}_{Q,\dset{z}}\subseteq \reals$, is such that for all $\lambda \in \set{A}_{Q, \dset{z}}$,
\begin{IEEEeqnarray}{rCl}
\label{EqReasonNisNormFoo}
\int \dot{f}^{-1}\left(-\frac{N_{Q, \dset{z}}(\lambda) + \foo{L}_{\dset{z}}(\thetav)}{\lambda}\right) \diff Q(\thetav) = 1.
\IEEEeqnarraynumspace
\end{IEEEeqnarray}
\end{subequations}
\end{definition}
The set $\set{A}_{Q, \dset{z}}$ in~\eqref{EqDefMapNormFunction} contains all the regularization factors for which Assumption \ref{assum:b} is satisfied. More specifically, it contains the regularization factors $\lambda$ for which the problem in~\eqref{EqOp_f_ERMRERNormal} has a solution.
Furthermore, the equality in~\eqref{EqReasonNisNormFoo} justifies referring to the function $N_{Q, \dset{z}}$ as the \emph{normalization function}, as it ensures that the measure $\Pgibbs{P}{Q}$ in~\eqref{EqGenpdffDv} is a probability measure.
 
 This section ends by highlighting that the probability measures $\Pgibbs{P}{Q}$ and $Q$ in~\eqref{EqGenpdffDv} are mutually absolutely continuous \cite[Corollary~1]{Perlaza-ISIT2024a}.

%
%
\section[ERM-fDR Dual Problem]{ERM-$f$DR Dual Problem}
\label{sec:dual}
The duality principle \cite[Chapter~5]{boyd2004convex} enables the reformulation of the optimization problem in~\eqref{EqOp_f_ERMRERNormal} into an alternative form, known as the dual problem.
In this section, this dual formulation is derived using the Legendre-Fenchel transform~\cite{boyd2004convex}, which is defined below.
\begin{definition}[Legendre-Fenchel transform {\cite{boyd2004convex}}]
\label{DefLT_cnvxcnj}
Consider a function $f:\set{I}\rightarrow \reals$, with $\set{I} \subset \reals$. The Legendre-Fenchel transform of the function $f$, denoted by $f^{*}:\set{J} \rightarrow \reals$, is 
\begin{IEEEeqnarray}{rCl}
\label{EqDefLT_cnvxcnj}
	f^{*}(t) & = & \sup_{x\in \set{I}}( tx- f(x)),
	\IEEEeqnarraynumspace
\end{IEEEeqnarray}
with
\vspace{-1mm}
\begin{IEEEeqnarray}{rCl}
\label{EqDefJinLFT}
	\set{J} & = & \{t \in \reals :f^{*}(t)<\infty\}.
	\IEEEeqnarraynumspace
\end{IEEEeqnarray}
\end{definition}
Using this notation, consider the following problem 
\begin{IEEEeqnarray}{rcl}
\label{EqOp_f_ERMRERDual}
  \min_{\beta \in \reals} & \quad \lambda \int f^{*}(-\frac{\beta+\foo{L}_{\dset{z}}(\thetav)}{\lambda})\diff Q(\thetav) + \beta,
\end{IEEEeqnarray}
where the real $\lambda$, the measure $Q$ and the function $f$ are those in~\eqref{EqOp_f_ERMRERNormal}; and the functions $\foo{L}_{\dset{z}}$ and $f^{*}$ are defined in~\eqref{EqLxy} and~\eqref{EqDefLT_cnvxcnj}, respectively.
The following theorem introduces the solution to the problem in~\eqref{EqOp_f_ERMRERDual}. 
\begin{theorem}
\label{Theo_dual_is_N}
Under Assumptions \ref{assum:a} and \ref{assum:b} the solution to the optimization problem in~\eqref{EqOp_f_ERMRERDual} is $N_{Q,\dset{z}}(\lambda)$, where the function $N_{Q,\dset{z}}$ is defined in~\eqref{EqDefNormFunction}.
\end{theorem}
\begin{IEEEproof}
Let $G:\reals \to \reals$ be a function such that 
\begin{IEEEeqnarray}{rcl}
\label{EqDefDualCost}
  	G(\beta) &\ = \ & \lambda \int f^{*}(-\frac{\beta+\foo{L}_{\dset{z}}(\thetav)}{\lambda})\diff Q(\thetav) + \beta,
\end{IEEEeqnarray}
which is the objective function of the optimization problem in~\eqref{EqOp_f_ERMRERDual}. 
Note that $G$ in~\eqref{EqDefDualCost} is a convex function, and thus satisfies:
\begin{IEEEeqnarray}{rcl}
  	\!\!\!\frac{\diff}{\diff \beta}G(\beta) 
        &\ = \ &\! \frac{\diff}{\diff \beta}(\lambda \int f^{*}(-\frac{\beta+\foo{L}_{\dset{z}}(\thetav)}{\lambda})\diff Q(\thetav) + \beta)\ \ \\
  	&\ = \ & \lambda \int \frac{\diff}{\diff \beta}f^{*}(-\frac{\beta+\foo{L}_{\dset{z}}(\thetav)}{\lambda})\diff Q(\thetav) + 1\\
  	&\ = \ & - \int \dot{f^{*}}(-\frac{\beta+\foo{L}_{\dset{z}}(\thetav)}{\lambda})\diff Q(\thetav) + 1,\label{EqDiffG_beta_s3}
\end{IEEEeqnarray}
where $\dot{f^{*}}$ is the derivative of the function $f^{*}$ in \eqref{EqOp_f_ERMRERDual}.
Let the solution to the optimization problem in~\eqref{EqDefDualCost} be denoted by $\widehat{\beta} \in \reals$ and note that the derivative of the function $G$ evaluated at $\widehat{\beta}$ is equal to zero, that is
\begin{IEEEeqnarray}{rcl}
  	\int \dot{f^{*}}(-\frac{\widehat{\beta}+\foo{L}_{\dset{z}}(\thetav)}{\lambda})\diff Q(\thetav) &\ = \ & 1.\label{EqDiffG_betaStar}
\end{IEEEeqnarray}
From~\cite[Corollary 23.5.1]{rockafellar1970conjugate} and Assumption~\ref{assum:a}, the following equality holds for all $t \in \set{J}$, with $\set{J}$ in~\eqref{EqDefJinLFT},
\begin{IEEEeqnarray}{rCCCl}
\label{EqDefLFTDiffF}
\frac{\diff}{\diff t}f^{*}(t) & = & \dot{f^{*}}(t)& = & \dot{f}^{-1}(t),
\end{IEEEeqnarray}
where the functions $\dot{f}^{-1}$ and $\dot{f^{*}}$ are defined in~\eqref{EqDefInvDiffF} and \eqref{EqDiffG_beta_s3}, respectively.
From~\eqref{EqDiffG_betaStar} and~\eqref{EqDefLFTDiffF}, it follows that
\begin{IEEEeqnarray}{rcl}
  	\int \dot{f}^{-1}(-\frac{\widehat{\beta}+\foo{L}_{\dset{z}}(\thetav)}{\lambda})\diff Q(\thetav) &\ = \ & 1,
\end{IEEEeqnarray}
which combined with~\eqref{EqReasonNisNormFoo} and Assumption~\ref{assum:b} yields
\begin{IEEEeqnarray}{rcl}
  	 N_{Q,\dset{z}}(\lambda) &\ = \ & \widehat{\beta},
\end{IEEEeqnarray}
and completes the proof.
\end{IEEEproof}

The following lemma establishes that the problem in~\eqref{EqOp_f_ERMRERDual} is the dual problem to the ERM-$f$DR problem in~\eqref{EqOp_f_ERMRERNormal} and characterizes the difference between their optimal values, which is often referred to as duality gap \cite[Section 8.3]{luenberger1964observing}.
\begin{lemma}
Under Assumptions~\ref{assum:a} and~\ref{assum:b}, the optimization problem in~\eqref{EqOp_f_ERMRERDual} is the dual problem to the ERM-$f$DR problem in~\eqref{EqOp_f_ERMRERNormal}.
Moreover, the duality gap is zero. 	
\end{lemma}
\begin{IEEEproof}
	Under Assumption~\ref{assum:a} and \cite[Section 3.3.2]{boyd2004convex}, it can be verified that for all $t \in \set{J}$, with $\set{J}$ in~\eqref{EqDefJinLFT}, the function $f^{*}$ in~\eqref{EqDefLT_cnvxcnj} satisfies 
	\begin{IEEEeqnarray}{rcl}
	\label{EqLFT_cool_equality}
  	f^{*}(t) &\ = \ & t\dot{f^{*}}(t)-f\big(\dot{f^{*}}(t)\big),
  	\IEEEeqnarraynumspace
	\end{IEEEeqnarray}
	where the function $\dot{f}^{*}$ is the same as in~\eqref{EqDefLFTDiffF}.
	From Assumption~\ref{assum:a} and~\eqref{EqDefLFTDiffF}, the Radon-Nikodym derivative $\frac{\diff \Pgibbs{P}{Q}}{\diff Q}$ in~\eqref{EqGenpdffDv} satisfies for all $\thetav \in \supp Q$,
	\begin{IEEEeqnarray}{rCl}
	\label{EqGenpdffDvLFT}
  	\frac{\diff \Pgibbs{P}{Q}}{\diff Q}(\thetav) &\ = \ & \dot{f^{*}}(-\frac{N_{Q, \dset{z}}(\lambda) + \foo{L}_{\dset{z}}(\thetav)}{\lambda}),
  	\IEEEeqnarraynumspace
	\end{IEEEeqnarray}
	where the functions $\foo{L}_{\dset{z}}$ and $N_{Q, \dset{z}}$ are defined in~\eqref{EqLxy} and \eqref{EqDefNormFunction}, respectively.
	Then, from~\eqref{EqLFT_cool_equality} and~\eqref{EqGenpdffDvLFT}, for all $\thetav \in \supp Q$, it holds that
	\begin{IEEEeqnarray}{rcl}
	\IEEEeqnarraymulticol{3}{l}{
  	\!\!\foo{L}_{\dset{z}}(\thetav)+\lambda f\Bigg(\frac{\diff \Pgibbs{P}{Q}}{\diff Q}(\thetav)\!\!\Bigg)\frac{\diff Q}{\diff \Pgibbs{P}{Q}}(\thetav)  
  	}\nonumber \\
  	\!& = \ & -\lambda f^{*}\!\Big(\!-\frac{ N_{Q, \dset{z}}(\lambda) + \foo{L}_{\dset{z}}(\thetav)}{\lambda}\Big)\frac{\diff Q}{\diff \Pgibbs{P}{Q}}(\thetav)\!-\! N_{Q, \dset{z}}(\lambda).  \ \ \ \ \label{EqPfDualLTArrange}
	\end{IEEEeqnarray}
	Taking the expectation in both sides of~\eqref{EqPfDualLTArrange} with respect to the probability measure $\Pgibbs{P}{Q}$ in~\eqref{EqGenpdffDv} yields
	\begin{IEEEeqnarray}{rcl}
	\IEEEeqnarraymulticol{3}{l}{
  	\foo{R}_{\dset{z}}(\Pgibbs{P}{Q})+\lambda \Divf{\Pgibbs{P}{Q}}{Q}  
  	}\nonumber \\
  	&\ = \ &\!- \lambda \int f^{*}\Big(-\frac{ N_{Q, \dset{z}}(\lambda) + \foo{L}_{\dset{z}}(\thetav)}{\lambda}\Big)\diff Q(\thetav)- N_{Q, \dset{z}}(\lambda).  \ \ \ \ \label{EqPfDualLTExpect}
	\end{IEEEeqnarray}
	Using Theorem~\ref{Theo_f_ERMRadNik} and Theorem~\ref{Theo_dual_is_N} in the left-hand and right-hand sides of~\eqref{EqPfDualLTExpect}, respectively, yields
	\begin{IEEEeqnarray}{rcl}
	\IEEEeqnarraymulticol{3}{l}{
  	\min_{P\in\bigtriangleup_{Q}(\set{M})}\foo{R}_{\dset{z}}(P)+\lambda \Divf{P}{Q}  
  	}\nonumber \\
  	&\ = \ & \max_{\beta\in \reals}- \lambda\! \int f^{*}\Big(-\frac{ \beta + \foo{L}_{\dset{z}}(\thetav)}{\lambda}\Big)\diff Q(\thetav)-\beta.  \ \  \label{EqPfDualLTOpt}
	\end{IEEEeqnarray}
	The proof that the optimization problem in~\eqref{EqOp_f_ERMRERDual} is the dual to the ERM-$f$DR problem in~\eqref{EqOp_f_ERMRERNormal} follows from~\eqref{EqPfDualLTOpt} and \cite[Theorem 1, Section 8.4]{luenberger1997bookOptimization}. The zero duality gap is established by the equality in~\eqref{EqPfDualLTOpt}, which completes the proof.
	\end{IEEEproof}

%
%
\section{Analysis of Regularization Factor}
\label{sec:analysisRegFact}
The purpose of this section is to characterize the function $N_{Q, \dset{z}}$ and the set $\set{A}_{Q, \dset{z}}$ in~\eqref{EqDefNormFunction}.
Given a real~$\delta\in [0, \infty)$, consider the Rashomon set $\set{L}_{\dset{z}}(\delta)$
, which is defined as follows
\begin{equation}
\label{EqType2LsetLamb2zero}
	\set{L}_{\dset{z}}(\delta) \triangleq \{\thetav \in \set{M}: \foo{L}_{\dset{z}}(\thetav) \leq \delta \}.
\end{equation}
Consider also the real numbers $\delta^\star_{Q, \dset{z}}$ and $\lambda^\star_{Q, \dset{z}}$ defined as follows
\vspace{-2mm}%
\begin{IEEEeqnarray}{rCl}
\label{EqDefDeltaStar}
\delta^\star_{Q, \dset{z}} &\triangleq & \inf \{\delta \in [0, \infty): Q(\set{L}_{\dset{z}}(\delta))>0\},
\end{IEEEeqnarray}
and
\vspace{-2mm}
\begin{IEEEeqnarray}{rCl}
	 \label{EqDefLambdaStar}
		\lambda^{\star}_{Q, \dset{z}} &\triangleq & \inf\set{A}_{Q,\dset{z}}.
\end{IEEEeqnarray}
Using this notation, the following theorem introduces one of the main properties of the function $N_{Q, \dset{z}}$.

\begin{theorem}
\label{theo_InfDevKfDR}
The function~$N_{Q, \dset{z}}$ in~\eqref{EqDefNormFunction} is strictly increasing and continuous within the interior of $\set{A}_{Q, \dset{z}}$ in~\eqref{EqDefMapNormFunction}. Furthermore,  for all $\lambda \in \set{A}_{Q, \dset{z}}$,
\begin{IEEEeqnarray}{rcl}
	N_{Q, \dset{z}}(\lambda) 
	&\ = \ & \lambda \frac{\diff }{\diff \lambda}N_{Q, \dset{z}}(\lambda) - \foo{R}_{\dset{z}}\Big(P^{(\lambda)}\Big),
\end{IEEEeqnarray}
where the probability measure $P^{(\lambda)} \in \bigtriangleup_{Q}(\set{M})$ satisfies for all $\thetav \in \supp Q$,
\begin{IEEEeqnarray}{rcl}
	\frac{\diff P^{(\lambda)}}{\diff Q}(\thetav) &\ = \ & \frac{\Bigg(\displaystyle\ddot{f}\Bigg(\frac{\diff \Pgibbs[\dset{z}]{P}{Q}}{\diff Q}(\thetav)\Bigg)\Bigg)^{-1}}{\displaystyle\int \Bigg(\ddot{f}\Bigg(\frac{\diff \Pgibbs[\dset{z}]{P}{Q}}{\diff Q}(\nuv)\Bigg)\Bigg)^{-1} \diff Q (\nuv)}.
\end{IEEEeqnarray}
\end{theorem}
\begin{IEEEproof}
	The proof is presented in Appendix~\ref{AppProofLemmaInfDevKDivf}.
\end{IEEEproof}
 
The continuity and monotonicity exhibited by the function $N_{Q,\dset{z}}$ allow the following characterization of the set $\set{A}_{Q, \dset{z}}$.
%
\begin{lemma}
\label{lemm_fDR_kset}
The set $\set{A}_{Q, \dset{z}}$ in~\eqref{EqDefMapNormFunction} is either empty or an interval that satisfies 
\begin{IEEEeqnarray}{rCl}
	\label{Eq_DivfConstrainOpen}
		\set{A}_{Q, \dset{z}} & = & 
		\begin{cases}
 			[\lambda^{\star}_{Q,\dset{z}}, \infty) & \text{if } \displaystyle \int\! \dot{f}^{-1}(\!- \frac{t +\foo{L}_{\dset{z}}(\thetav)}{\lambda^{\star}_{Q,\dset{z}}})\diff Q(\thetav) < \infty,\vspace{2mm} \\
 			(\lambda^{\star}_{Q,\dset{z}}, \infty)& \text{otherwise},
 		\end{cases}
	\end{IEEEeqnarray}
where $t > \lim_{\lambda \rightarrow {\lambda^{\star}_{Q, \dset{z}}}^{+}} N_{Q, \dset{z}}(\lambda)$ and $N_{Q, \dset{z}}$ is defined in~\eqref{EqDefNormFunction}.
\end{lemma}
\begin{IEEEproof}
	The proof is presented in Appendix~\ref{app_proof_lemm_fDR_kset}.
\end{IEEEproof}

Lemma~\ref{lemm_fDR_kset} highlights two facts. First, the set $\set{A}_{Q, \dset{z}}$ is a convex subset of positive reals. 
Second, if there exists a solution to the optimization problem in~\eqref{EqOp_f_ERMRERNormal} for some $\lambda>0$, then there exists a solution to such a problem when $\lambda$ is replaced by~$\bar{\lambda} \in(\lambda,\infty)$.

The following lemma presents a case in which the set $\set{A}_{Q, \dset{z}}$ in~\eqref{EqDefMapNormFunction} can be fully characterized.
\begin{lemma}
\label{lemm_fDR_No_minRegF_nneg}
If the function $\dot{f}^{-1}$ in~\eqref{EqGenpdffDv} is strictly positive, then the set $\set{A}_{Q, \dset{z}}$ is identical to $(0,\infty)$.
\end{lemma}
\begin{IEEEproof}
	The proof is presented in Appendix~\ref{app_proof_lemm_fDR_No_minRegF_nneg}.
\end{IEEEproof}

\section{Exact Characterization of the Generalization Error}
\label{sec:GenErr}
Let the functional $\foo{G}:(\set{X}\times\set{Y})^n\times\bigtriangleup(\set{M})\times\bigtriangleup(\set{M})\to \reals$ satisfy
\vspace{-2mm}
\begin{IEEEeqnarray}{rCl}
\label{EqGap}
  \foo{G}(\dset{z},P_1,P_2)
   & = & \foo{R}_{\dset{z}}(P_1)- \foo{R}_{\dset{z}}(P_2).
\end{IEEEeqnarray}
The value $\foo{G}(\dset{z},P_1,P_2)$ in~\eqref{EqGap} represents the variation of the functional $\foo{R}_{\dset{z}}$ in~\eqref{EqRxy} when its argument changes from $P_2$ to $P_1$.
This variation is referred to as an \emph{algorithm driven gap} in~\cite{perlaza2024Generalization}, which is justified by the fact that $P_1$ and $P_2$ can be assimilated to learning algorithms.
Using this notion, this section studies the generalization error of machine learning algorithms. 
\begin{definition}[Generalization Error {\cite[Definition 4]{perlaza2024Generalization}}]
\label{DefGenError}
The generalization error induced by the algorithm $P_{\Thetam|\dset{Z}} \in \bigtriangleup(\set{M}|(\set{X}\times\set{Y})^n)$ under the assumption that training and test datasets are independently sampled from a probability measure~$P_{\vect{Z}} \in \triangle\left( \left(\set{X} \times \set{Y}\right)^n \right)$, which is denoted by $\bar{\bar{\foo{G}}}(P_{\Thetam|\vect{Z}},P_{\vect{Z}})$, is
\vspace{-2mm}
\begin{IEEEeqnarray}{rCl}
\IEEEeqnarraymulticol{3}{l}{
 \!\!\bar{\bar{\foo{G}}}(P_{\Thetam|\vect{Z}},P_{\vect{Z}})
}\nonumber\\ 
  \!\!& = &\! \int\!\!\! \int\!\! (\foo{R}_{\dset{u}}(P_{\Thetam|\vect{Z}=\dset{z}}) \!-\! \foo{R}_{\dset{z}}(P_{\Thetam|\vect{Z}=\dset{z}}))\! \diff P_{\vect{Z}}(\dset{u})\! \diff P_{\vect{Z}}(\dset{z}).\ \label{EqDefGenError}
\end{IEEEeqnarray}
\end{definition}
Consider the following assumptions:
\begin{itemize}
\item[\namedlabel{assum:d}{$(d)$}] For all $\dset{z} \in (\set{X}\times\set{Y})^n$, the probability measure $P_{\Thetam|\dset{Z}=\dset{z}}$ is absolutely continuous with respect to the probability measure $P_{\Thetam}\in \bigtriangleup_{Q}(\set{M})$, which satisfies for all measurable subsets $\set{C}$ of $\set{M}$
\vspace{-1mm}
\begin{IEEEeqnarray}{rCl}
\label{EqDefPTheta}
	P_{\Thetam}(\set{C}) & = & \int P_{\Thetam|\dset{Z}=\dset{z}}(\set{C})\diff P_{\dset{Z}}(\dset{z}).
\end{IEEEeqnarray}

\item[\namedlabel{assum:e}{$(e)$}] The probability measure $P_{\Thetam} $ in~\eqref{EqDefPTheta} and $Q$ in~\eqref{EqOp_f_ERMRERNormal} are mutually absolutely continuous.
\end{itemize}
Under Assumptions~\ref{assum:d} and \ref{assum:e}, it follows from~\cite[Lemma 3]{perlaza2024Generalization} that the generalization error $\bar{\bar{\foo{G}}}(P_{\Thetam|\dset{Z}},P_{\dset{Z}})$ in~\eqref{EqDefGenError} satisfies
\begin{IEEEeqnarray}{rCl}
\IEEEeqnarraymulticol{3}{l}{
\!\!\!\bar{\bar{\foo{G}}}(P_{\Thetam|\dset{Z}},P_{\dset{Z}})
}\nonumber \\
	\!\!\!& = &\!\! \int\!\! \foo{G}(\dset{z}, P_{\Thetam}, P_{\Thetam|\dset{Z}=\dset{z}})\diff P_{\dset{Z}}(\dset{z})\\
	\!\!\!& = &\!\! \int\!\! \foo{G}(\!\dset{z},P_{\Thetam}, \Pgibbs{P}{Q}\!\!)\!\!-\!\foo{G}(\!\dset{z}, P_{\Thetam|\dset{Z}=\dset{z}},\Pgibbs{P}{Q}\!\!)\!\diff \! P_{\dset{Z}}(\dset{z}),\ \ \ \label{EqGenDiffPtoP}
\end{IEEEeqnarray}
where the measures $\Pgibbs{P}{Q}$ and $P_{\Thetam}$ are defined in~\eqref{EqGenpdffDv} and~\eqref{EqDefPTheta}, respectively; and the functional $\foo{G}$ is defined in~\eqref{EqGap}. 
The following theorem presents the main tool in this section.
%
\begin{theorem}
\label{theo_ERM_fDR_LT_AnyP}
The probability measure $\Pgibbs{P}{Q}$ in~\eqref{EqGenpdffDv} satisfies for all $P \in \bigtriangleup_{Q}(\set{M})$,
\begin{IEEEeqnarray}{rCl}
        \!\!\!\foo{G}\Big(\!\dset{z},P,\Pgibbs{P}{Q}\!\Big)
	\!& = &\! \lambda \!\!\int \!\Bigg(\!1-\frac{\diff P}{\diff \Pgibbs{P}{Q}}(\thetav)\!\!\Bigg)\!
        \Bigg(\! f\Bigg(\!\frac{\diff \Pgibbs{P}{Q}}{\diff Q}(\thetav)\!\Bigg)  
	\nonumber \\ & &
	+ f^{*}(-\frac{\foo{L}_{\dset{z}}(\thetav) + N_{Q,\dset{z}}(\lambda)}{\lambda})\Bigg)
	\diff Q(\thetav),\quad \label{EqTheo_ERM_fDR_LT}
\end{IEEEeqnarray}
where the functions $\foo{L}_{\dset{z}}$, $N_{Q,\dset{z}}$, and $f^{*}$ are defined in~\eqref{EqLxy},~\eqref{EqDefNormFunction} and~\eqref{EqDefLT_cnvxcnj}, respectively; and the Radon-Nikodym derivative $\frac{\diff \Pgibbs{P}{Q}}{\diff Q}$ is defined in~\eqref{EqGenpdffDv}.
\end{theorem}
\begin{IEEEproof}
	The proof is presented in Appendix~\ref{app_theo_ERM_fDR_LT_AnyP}.
\end{IEEEproof}
Theorem~\ref{theo_ERM_fDR_LT_AnyP} allows generalizing the method of algorithm-driven gaps introduced in \cite{perlaza2024Generalization}. In particular the choice of $f(x)=x\log(x)$ in Theorem~\ref{theo_ERM_fDR_LT_AnyP} leads to \cite[Theorem~1]{Perlaza-ISIT2023b}. 
Moreover, using Theorem~\ref{theo_ERM_fDR_LT_AnyP} in~\eqref{EqGenDiffPtoP} leads to the following characterization of $\bar{\bar{\foo{G}}}(P_{\Thetam|\vect{Z}},P_{\vect{Z}})$ in~\eqref{EqDefGenError}. 

\begin{theorem}
\label{Theo_ERM_fDR_LT_GenERr}
The generalization error $\bar{\bar{\foo{G}}}(P_{\Thetam|\vect{Z}},P_{\vect{Z}})$ in~\eqref{EqDefGenError}, under Assumptions~\ref{assum:a}, \ref{assum:b}, \ref{assum:d} and \ref{assum:e}, satisfies 
\begin{IEEEeqnarray}{rCl}
\IEEEeqnarraymulticol{3}{l}{
	\bar{\bar{\foo{G}}}(P_{\Thetam|\vect{Z}},P_{\vect{Z}})
	}\nonumber \\
	& = & \lambda \!\int\!\int\! \Bigg(f\Bigg(\frac{\diff \Pgibbs{P}{Q}}{\diff Q}(\thetav)\Bigg)  
	\!+\! f^{*}\Bigg(\!-\frac{\foo{L}_{\dset{z}}(\thetav) + N_{Q,\dset{z}}(\lambda)}{\lambda}\Bigg)\Bigg)
	\nonumber \\ & &
	\Bigg(\frac{\diff P_{\Thetam}}{\diff \Pgibbs{P}{Q}}(\thetav)
    -\frac{\diff P_{\Thetam|\bm{Z}=\dset{z}}}{\diff \Pgibbs{P}{Q}}(\thetav)
    \Bigg)\diff Q(\thetav) \diff P_{\dset{Z}}(\dset{z}),\quad \label{EqTheo_ERM_fDR_GE}
\end{IEEEeqnarray}
where the functions $\foo{L}_{\dset{z}}$, $N_{Q,\dset{z}}$, and $f^{*}$ are defined in~\eqref{EqLxy},~\eqref{EqDefNormFunction} and~\eqref{EqDefLT_cnvxcnj}, respectively; and the probability measure $\Pgibbs{P}{Q}$ and the real $\lambda$ are those in~\eqref{EqGenpdffDv}.
\end{theorem}
\begin{IEEEproof}
	The proof is presented in Appendix~\ref{app_theo_ERM_fDR_LT_GenERr}.
\end{IEEEproof}

The expression in~\eqref{EqTheo_ERM_fDR_GE} significantly simplifies for some choices of $f$. See for instance, the case in which $f(x) = x\log(x)$ in \cite[Lemma 4]{perlaza2024Generalization}. Another case in which such expression becomes particularly simple is the case in which the algorithm $P_{\vect{\Theta} | \vect{Z}}$ is the solution to the optimization problems~\eqref{EqOp_f_ERMRERNormal} and~\eqref{EqOp_f_ERM_RND2}. In such a case, the following holds.
\begin{theorem}\label{theo_ERM_fDR_gen}
Consider the solution to the optimization problem in~\eqref{EqOp_f_ERMRERNormal}, $\Pgibbs{P}{Q}$ in~\eqref{EqGenpdffDv}, and consider also the generalization error  $\bar{\bar{\foo{G}}}\big(P^{(Q,\lambda)}_{\Thetam|\vect{Z}},P_{\vect{Z}}\big)$ defined as in~\eqref{EqDefGenError}. Under Assumptions \ref{assum:a}, \ref{assum:b},  \ref{assum:d} and \ref{assum:e}, the following holds
\begin{IEEEeqnarray}{rCl}
    \!\bar{\bar{\foo{G}}}(\!P^{(Q,\lambda)}_{\Thetam|\vect{Z}}\!, P_{\vect{Z}}\!)\!
    \!& = &\lambda\! \int\!\Bigg(\int\dot{f}\Bigg(\frac{\diff \Pgibbs{P}{Q}}{\diff Q} ( \thetav ) \Bigg)\diff \Pgibbs{P}{Q}(\thetav) 
    \nonumber \\ &  &
    \!\!\!\!\!\!-\!\!\int\!\! \dot{f}\Bigg(\!\frac{\diff \Pgibbs{P}{Q}}{\diff Q} (\thetav)\!\! \Bigg)\!\diff \PgibbsNonCond{P}{Q}\!(\thetav)\!\!\Bigg) \!\!\diff P_{\vect{Z}} (\vect{z})\!,\ \ \label{EqGenExact}
\end{IEEEeqnarray}
where the function $\dot{f}$ is defined in~\eqref{EqDefDiffF}. 
\end{theorem}
\begin{IEEEproof}
	The proof is presented in \cite[Appendix D-C]{daunas2024Arxiv}.
\end{IEEEproof}
%
	The case in which $f(x) = x\log(x)$ in Theorem~\ref{Theo_ERM_fDR_LT_GenERr} and Theorem~\ref{theo_ERM_fDR_gen}, which corresponds to the Gibbs algorithm, leads to the existing results in~\cite[Theorem 16]{PerlazaTIT2024} and \cite[Theorem~1]{aminian2021exact}.

\section{Conclusions}\label{SecConclusions}
This work has established the conditions under which the solution to the ERM-$f$DR problem in~\eqref{EqOp_f_ERMRERNormal} also serves as the solution to the optimization problem in~\eqref{EqOp_f_ERM_RND2}.
Furthermore, the solution of the ERM-$f$DR problem has been connected to the Legendre-Fenchel transform by establishing the dual problem in~\eqref{EqOp_f_ERMRERDual} to the optimization problem in~\eqref{EqOp_f_ERMRERNormal}. This connection between the dual formulation and the Legendre-Fenchel transform, through the application of the implicit function theorem, provides an explicit expression for the normalization function.
Notably, the link to the Legendre-Fenchel transform, under mild assumptions, enables the derivation of a explicit expression for the generalization error of general learning algorithms. Lastly, it offers a separate explicit expression for evaluating the generalization error of algorithms arising as the solution to an ERM-$f$DR optimization problem.

%
\newpage
\IEEEtriggeratref{17}
\bibliographystyle{IEEEtranlink}
\bibliography{iEEEtranBibStyle.bib}

\begin{thebibliography}{10}
\providecommand{\url}[1]{#1}
\csname url@samestyle\endcsname
\providecommand{\newblock}{\relax}
\providecommand{\bibinfo}[2]{#2}
\providecommand{\BIBentrySTDinterwordspacing}{\spaceskip=0pt\relax}
\providecommand{\BIBentryALTinterwordstretchfactor}{4}
\providecommand{\BIBentryALTinterwordspacing}{\spaceskip=\fontdimen2\font plus
\BIBentryALTinterwordstretchfactor\fontdimen3\font minus \fontdimen4\font\relax}
\providecommand{\BIBforeignlanguage}[2]{{%
\expandafter\ifx\csname l@#1\endcsname\relax
\typeout{** WARNING: IEEEtranlink.bst: No hyphenation pattern has been}%
\typeout{** loaded for the language `#1'. Using the pattern for}%
\typeout{** the default language instead.}%
\else
\language=\csname l@#1\endcsname
\fi
#2}}
\providecommand{\BIBdecl}{\relax}
\BIBdecl

\bibitem{vapnik1964perceptron}
V.~Vapnik and A.~Y. Chervonenkis, ``On a perceptron class,'' \emph{Avtomatika i Telemkhanika}, vol.~25, no.~1, pp. 112--120, Feb. 1964.

\bibitem{vapnik1992principles}
V.~Vapnik, ``Principles of risk minimization for learning theory,'' \emph{Advances in Neural Information Processing Systems}, vol.~4, pp. 831--838, Jan. 1992.

\bibitem{raginsky2016information}
M.~Raginsky, A.~Rakhlin, M.~Tsao, Y.~Wu, and A.~Xu, ``Information-theoretic analysis of stability and bias of learning algorithms,'' in \emph{Proceedings of the IEEE Information Theory Workshop (ITW)}, Cambridge, UK, Sep. 2016, pp. 26--30.

\bibitem{russo2019much}
D.~Russo and J.~Zou, ``How much does your data exploration overfit? {C}ontrolling bias via information usage,'' \emph{IEEE Transactions on Information Theory}, vol.~66, no.~1, pp. 302--323, Jan. 2019.

\bibitem{perlaza2024ERMRER}
\BIBentryALTinterwordspacing
S.~M. Perlaza, G.~Bisson, I.~Esnaola, A.~Jean-Marie, and S.~Rini, ``\href{https://doi.org/10.1109/TIT.2024.3365728}{Empirical risk minimization with relative entropy regularization},'' \emph{IEEE Transactions on Information Theory}, vol.~70, no.~7, pp. 5122 -- 5161, Jul. 2024.

\bibitem{Perlaza_ISIT_2022}
\BIBentryALTinterwordspacing
------, ``\href{https://doi.org/10.1109/ISIT50566.2022.9834273}{Empirical risk minimization with relative entropy regularization: {O}ptimality and sensitivity},'' in \emph{Proceedings of the IEEE International Symposium on Information Theory (ISIT)}, Espoo, Finland, Jul. 2022, pp. 684--689.

\bibitem{renyi1961measures}
A.~R{\'e}nyi, ``On measures of information and entropy,'' in \emph{Proceedings of the 4th Berkeley Symposium on Mathematics, Statistics and Probability}, Berkeley, CA, USA, Jun. 1961, pp. 547--561.

\bibitem{sason2016fdivergence}
I.~Sason and S.~Verdú, ``$f$-divergence inequalities,'' \emph{IEEE Transactions on Information Theory}, vol.~62, no.~11, pp. 5973--6006, Jun. 2016.

\bibitem{csiszar1967information}
I.~Csisz{\'a}r, ``Information-type measures of difference of probability distributions and indirect observation,'' \emph{Studia Scientiarum Mathematicarum Hungarica}, vol.~2, no.~1, pp. 299--318, Jun. 1967.

\bibitem{teboulle1992entropic}
M.~Teboulle, ``Entropic proximal mappings with applications to nonlinear programming,'' \emph{Mathematics of Operations Research}, vol.~17, no.~3, pp. 670--690, Aug. 1992.

\bibitem{beck2003mirror}
A.~Beck and M.~Teboulle, ``Mirror descent and nonlinear projected subgradient methods for convex optimization,'' \emph{Operations Research Letters}, vol.~31, no.~3, pp. 167--175, Jan. 2003.

\bibitem{alquier2021non}
P.~Alquier, ``Non-exponentially weighted aggregation: regret bounds for unbounded loss functions,'' in \emph{Proceedings of the 38th International Conference on Machine Learning (ICML)}, vol. 139, Jul. 2021, pp. 207--218.

\bibitem{Perlaza-ISIT2024a}
\BIBentryALTinterwordspacing
F.~Daunas, I.~Esnaola, S.~M. Perlaza, and H.~V. Poor, ``\href{https://doi.org/10.1109/ISIT57864.2024.10619260}{Equivalence of empirical risk minimization to regularization on the family of f-divergences},'' in \emph{Proceedings of the IEEE International Symposium on Information Theory (ISIT)}, Athens, Greece, Jul. 2024.

\bibitem{daunas2024TITAsymmetry}
\BIBentryALTinterwordspacing
------, ``\href{https://inria.hal.science/hal-04789606/document}{Asymmetry of the relative entropy in the regularization of empirical risk minimization},'' \emph{Submitted to IEEE Transactions on Information Theory}, pp. 1--1, Oct. 2024.

\bibitem{Perlaza-ISIT2023a}
\BIBentryALTinterwordspacing
------, ``\href{https://doi.org/10.1109/ISIT54713.2023.10206876}{Analysis of the relative entropy asymmetry in the regularization of empirical risk minimization},'' in \emph{Proceedings of the IEEE International Symposium on Information Theory (ISIT)}, Taipei, Taiwan, Jun. 2023.

\bibitem{perlaza2024Generalization}
\BIBentryALTinterwordspacing
S.~M. Perlaza and X.~Zou, ``\href{https://inria.hal.science/hal-04789606/document}{The generalization error of machine learning algorithms},'' \emph{Submitted to IEEE Transactions on Information Theory}, pp. 1--1, Nov. 2024.

\bibitem{zou2024WorstCase}
\BIBentryALTinterwordspacing
X.~Zou, S.~M. Perlaza, I.~Esnaola, E.~Altman, and H.~V. Poor, ``\href{https://doi.org/10.1109/JSAIT.2024.3383281}{The worst-case data-generating probability measure in statistical learning},'' \emph{IEEE Journal on Selected Areas in Information Theory}, vol.~5, no.~1, pp. 175 -- 189, Apr. 2024.

\bibitem{zou2024generalization}
\BIBentryALTinterwordspacing
X.~Zou, S.~M. Perlaza, I.~Esnaola, and E.~Altman, ``\href{https://doi.org/10.1609/aaai.v38i15.29674}{Generalization analysis of machine learning algorithms via the worst-case data-generating probability measure},'' in \emph{Proceedings of the AAAI Conference on Artificial Intelligence}, Vancouver, Canada, Feb. 2024, pp. 17\,271--17\,279.

\bibitem{bulatov2005complexity}
\BIBentryALTinterwordspacing
A.~Bulatov and M.~Grohe, ``\href{https://doi.org/10.1016/j.tcs.2005.09.011}{The complexity of partition functions},'' \emph{Theoretical Computer Science}, vol. 348, no.~2, pp. 148--186, Sep. 2005.

\bibitem{kullback1951information}
S.~Kullback and R.~A. Leibler, ``On information and sufficiency,'' \emph{Annals of Mathematical Statistics}, vol.~22, no.~1, pp. 79--86, Mar. 1951.

\bibitem{rockafellar1970conjugate}
R.~T. Rockafellar, \emph{Conjugate Convex Functions in Optimal Control and the Calculus of Variations}, 2nd~ed.\hskip 1em plus 0.5em minus 0.4em\relax Princeton, NJ, USA: Princeton University Press, 1970.

\bibitem{boyd2004convex}
S.~Boyd, S.~P. Boyd, and L.~Vandenberghe, \emph{Convex optimization}, 1st~ed.\hskip 1em plus 0.5em minus 0.4em\relax Cambridge, UK: Cambridge University Press, 2004.

\bibitem{InriaRR9474}
\BIBentryALTinterwordspacing
S.~M. Perlaza, I.~Esnaola, G.~Bisson, and H.~V. Poor, ``\href{https://hal.science/hal-03703628v3}{Sensitivity of the {G}ibbs algorithm to data aggregation in supervised machine learning},'' INRIA, Centre Inria d'Universit\'e C\^ote d'Azur, Sophia Antipolis, France, Tech. Rep. RR-9474, Jun. 2022.

\bibitem{gateaux1913fonctionnelles}
R.~Gateaux, ``Sur les fonctionnelles continues et les fonctionnelles analytiques,'' \emph{Comptes rendus hebdomadaires des séances de l'{A}cadémie des {S}ciences, Paris}, vol. 157, no. 325-327, p.~65, 1913.

\bibitem{luenberger1997bookOptimization}
D.~G. Luenberger, \emph{Optimization by Vector Space Methods}, 1st~ed.\hskip 1em plus 0.5em minus 0.4em\relax New York, NY, USA: Wiley, 1997.

\bibitem{daunas2024Arxiv}
F.~Daunas, I.~Esnaola, S.~M. Perlaza, and G.~Aminian, ``Generalization error of f-divergence stabilized algorithms via duality,'' \emph{\normalfont{arXiv preprint arXiv:}}, Jan. 2025.

\bibitem{luenberger1964observing}
D.~G. Luenberger, ``Observing the state of a linear system,'' \emph{IEEE Transactions on Military Electronics}, vol.~8, no.~2, pp. 74--80, Apr. 1964.

\bibitem{Perlaza-ISIT2023b}
\BIBentryALTinterwordspacing
S.~M. Perlaza, I.~Esnaola, G.~Bisson, and H.~V. Poor, ``\href{https://ieeexplore.ieee.org/stamp/stamp.jsp?arnumber=10206506}{On the validation of {G}ibbs algorithms: {T}raining datasets, test datasets and their aggregation},'' in \emph{Proceedings of the IEEE International Symposium on Information Theory (ISIT)}, Taipei, Taiwan, Jun. 2023.

\bibitem{PerlazaTIT2024}
\BIBentryALTinterwordspacing
S.~M. Perlaza, G.~Bisson, I.~Esnaola, A.~Jean-Marie, and S.~Rini, ``\href{https://ieeexplore.ieee.org/stamp/stamp.jsp?arnumber=10433697}{Empirical risk minimization with relative entropy regularization},'' \emph{IEEE Transactions on Information Theory}, vol.~70, no.~7, pp. 5122 -- 5161, Jul. 2024.

\bibitem{aminian2021exact}
\BIBentryALTinterwordspacing
G.~Aminian, Y.~Bu, L.~Toni, M.~Rodrigues, and G.~Wornell, ``\href{https://proceedings.neurips.cc/paper_files/paper/2021/file/445e24b5f22cacb9d51a837c10e91a3f-Paper.pdf}{An exact characterization of the generalization error for the {G}ibbs algorithm},'' \emph{Advances in Neural Information Processing Systems}, vol.~34, pp. 8106--8118, Dec. 2021.

\bibitem{ash2000probability}
R.~B. Ash and C.~A. Doleans-Dade, \emph{Probability and Measure Theory}, 2nd~ed.\hskip 1em plus 0.5em minus 0.4em\relax Burlington, MA, USA: Academic Press, 2000.

\bibitem{bartle2000introduction}
R.~G. Bartle and D.~R. Sherbert, \emph{Introduction to Real Analysis}, 2nd~ed.\hskip 1em plus 0.5em minus 0.4em\relax New York, NY, USA: Wiley New York, 2000.

\bibitem{douchet2010Analyse}
J.~Douchet, \emph{Analyse : {R}ecueil d'Exercices et Aide-M\'emoire}, 3rd~ed.\hskip 1em plus 0.5em minus 0.4em\relax Lausanne, Switzerland: PPUR, 2010, vol.~1.

\bibitem{oswaldo2013TIFT}
\BIBentryALTinterwordspacing
O.~de~Oliveira, ``\href{https://projecteuclid.org/journals/real-analysis-exchange/volume-39/issue-1/The-Implicit-and-Inverse-Function-Theorems-Easy-Proofs/rae/1404230147.full}{{The Implicit and Inverse Function Theorems: Easy Proofs}},'' \emph{Real Analysis Exchange}, vol.~39, no.~1, pp. 207 -- 218, 2013.

\bibitem{rudin1953bookPrinciples}
W.~Rudin, \emph{Principles of Mathematical Analysis}, 1st~ed.\hskip 1em plus 0.5em minus 0.4em\relax New York, NY, USA: McGraw-Hill Book Company, Inc., 1953.

\end{thebibliography}
%

\newpage

\onecolumn
\appendices

\section{Preliminaries}
\begin{theorem}
\label{Theo_ERM_fDR_Leibniz}
Given a probability measure space $\mspc{\set{M}}{\mu}$	and an open subset $\set{A}$ of $\reals$, let the function $f: \set{A}\times\set{M} \rightarrow \reals$ be measurable with respect to $\msblspc{\set{A}\times\set{M}}$ and $\bormsblspc{\reals}$. If for all $\nu \in \set{M}$, the function $f(\cdot,\nuv):\set{A}_{Q, \dset{z}} \rightarrow \reals$ is \Lipschitz continuous and for some $u \in \set{A}$, $\int f(u,\nu) \diff \mu(\nu) < \infty$, then
\begin{IEEEeqnarray}{rCl}
\label{EqLeibnizInt}
	\left.\frac{\diff }{\diff t} \int f(t, \nu) \diff \mu(\nu)\right|_{t = u}
	& = & \left.\int \frac{\diff }{\diff t} f(t, \nu) \diff \mu(\nu)\right|_{t = u},
\end{IEEEeqnarray}
\end{theorem}
\begin{IEEEproof}
Note that
\begin{IEEEeqnarray}{rCl}
	\left.\frac{\diff }{\diff t} \int f(t, \nu) \diff \mu(\nu)\right|_{t = u}
	& = & \lim_{\delta \rightarrow 0} \frac{\int f(u+\delta, \nu) \diff \mu(\nu)-\int f(u, \nu) \diff \mu(\nu)}{\delta}
	\label{EqLeibnizLim_s1}\\
	& = & \lim_{\delta \rightarrow 0} \int \frac{f(u+\delta, \nu) - f(u, \nu) }{\delta} \diff \mu(\nu),
	\label{EqLeibnizLim_s2}
\end{IEEEeqnarray}
where~\eqref{EqLeibnizLim_s2} follows from Theorem 1.6.3 in~\cite{ash2000probability}. The assumption that for all $\nu \in \set{M}$, the function $f(\cdot,\nuv)$ is \Lipschitz continuous implies that for all $u \in \set{A}$ and some $\delta \in \reals$,
\begin{IEEEeqnarray}{rCl}
	\abs{f(u+\delta, \nu) - f(u, \nu)}
	& < & L \abs{\delta},
	\label{EqLeibnizIsLipschitz_s1}\\
\end{IEEEeqnarray}
with $L < \infty$. And thus, dividing the RHS and LHS of~\eqref{EqLeibnizIsLipschitz_s1} by $\abs{\delta}$ yields
\begin{IEEEeqnarray}{rCl}
	\abs{\frac{f(u+\delta, \nu) - f(u, \nu)}{\delta}}
	& < & L ,
	\label{EqLeibnizIsLipschitz_s2}\\
\end{IEEEeqnarray}
which implies that
\begin{IEEEeqnarray}{rCl}
	\int \abs{\frac{f(u+\delta, \nu) - f(u, \nu)}{\delta}} \diff \mu(\nu)
	& < & \infty.
	\label{EqLeibnizIsLipschitz_s3}\\
\end{IEEEeqnarray}
This allows using the dominated convergence theorem presented by~\cite{ash2000probability} in Theorem 1.6.9, as follows. From~\eqref{EqLeibnizLim_s2}, the following holds
\begin{IEEEeqnarray}{rCl}
	\left.\frac{\diff }{\diff t} \int f(t, \nu) \diff \mu(\nu)\right|_{t = u}
	& = & \lim_{\delta \rightarrow 0} \int \frac{ f(u+\delta, \nu) - f(u, \nu) }{\delta} \diff \mu(\nu)
	\label{EqLeibnizLim2_s1}\\
	& = & \int  \lim_{\delta \rightarrow 0} \frac{ f(u+\delta, \nu) - f(u, \nu) }{\delta} \diff \mu(\nu)
	\label{EqLeibnizLim2_s2}\\
	& = & \left.\int \frac{\diff }{\diff t} f(t, \nu) \diff \mu(\nu)\right|_{t = u},
\end{IEEEeqnarray}
where~\eqref{EqLeibnizLim2_s2} follows from the dominated convergence theorem, Theorem 1.6.9 in~\cite{ash2000probability}. This completes the proof.
\end{IEEEproof}

\begin{lemma}
\label{lemm_ERM_fDR_DiffZero}
Let $\field{M}$ be the set of measurable functions $h:\set{M} \rightarrow \reals$, with respect to the measurable space $\msblspc{\set{M}}$ and $\bormsblspc{\reals}$ and $h \in \field{M}$. Let $\field{S}$ be the subset of $\field{M}$, including all nonnegative functions that are absolutely integrable with respect to a probability measure $Q$. That is, for all $h \in \field{S}$, it holds that
\begin{IEEEeqnarray}{rCl}
\int \abs{h(\thetav)}\diff Q(\thetav) & < & \infty.
\end{IEEEeqnarray}
Given a strictly convex function $f:\reals \rightarrow \reals$, let the function $\hat{r}: \reals \rightarrow \reals$ be such that
\begin{IEEEeqnarray}{rCl}
\label{EqHatR_lemmAppx}
	\hat{r}(\alpha) & = & \int f(g(\thetav) + \alpha h(\thetav)) \diff Q(\thetav),
\end{IEEEeqnarray}
for some function $g$ and $h$ in $\field{S}$ and $\alpha \in (-\epsilon, \epsilon)$, with $\epsilon > 0$ arbitrarily small. Then, the function $\hat{r}$ in~\eqref{EqHatR_lemmAppx} is differentiable at zero.
\end{lemma}
\begin{IEEEproof}
The objective is to prove that the function $\hat{r}$ in~\eqref{EqHatR_lemmAppx} is differentiable at zero, which boils down to proving that the limit
\begin{IEEEeqnarray}{rCl}
\label{EqLimf}
	& \lim_{\delta \rightarrow 0} \frac{1}{\delta}(\hat{r}(\alpha+\delta) -\hat{r}(\alpha))&
\end{IEEEeqnarray}
exists for $\alpha \in (-\epsilon, \epsilon)$, with $\epsilon > 0$ arbitrarily small. 
The proof of the existence of such limit in~\eqref{EqLimf} relies on the fact that the function $f$ in~\eqref{EqHatR_lemmAppx} is strictly convex and differentiable, which implies that $f$ is also Lipschitz continuous. Hence, it follows that
\begin{IEEEeqnarray}{rCl}
\label{EqfisLipschitz}
	\abs{f(g(\thetav) + (\alpha+\delta) h(\thetav))-f(g(\thetav) + \alpha h(\thetav))} & \leq & c\abs{h(\thetav)}\abs{\delta},
\end{IEEEeqnarray}
for some positive and finite constant $c$, which implies that
\begin{IEEEeqnarray}{rCl}
\label{EqDfDxLipschitz}
	\frac{\abs{f(g(\thetav) + (\alpha+\delta) h(\thetav))-f(g(\thetav) + \alpha h(\thetav))}}{\abs{\delta}} & \leq & c\abs{h(\thetav)},
\end{IEEEeqnarray}
and thus, given that $g \in \field{S}$, it holds that
\begin{IEEEeqnarray}{rCl}
\label{EqLimitExists}
	\int \frac{\abs{f(g(\thetav) + (\alpha+\delta) h(\thetav))-f(g(\thetav) + \alpha h(\thetav))}}{\abs{\delta}} \diff Q(\thetav) & \leq & \infty.,
\end{IEEEeqnarray}
This allows using the dominated convergence theorem as follows. From the fact that the function $f$ is different, let $\dot{f}:(\infty,\infty) \rightarrow \reals$ be the first derivative of $f$. The limit in~\eqref{EqLimf} satisfies for $\alpha \in \alpha \in (-\epsilon,\epsilon)$, with $\epsilon > 0$ arbitrarily small,
\begin{IEEEeqnarray}{rCl}
	\lim_{\delta \rightarrow 0} \frac{1}{\delta}(\hat{r}(\alpha+\delta) -\hat{r}(\alpha)) 
	& = & \lim_{\delta \rightarrow 0} \frac{1}{\delta}(\int f(g(\thetav) + (\alpha+\delta) h(\thetav)) \diff Q(\thetav)-\int f(g(\thetav) + \alpha h(\thetav)) \diff Q(\thetav))
	\label{EqLimHatrDCT_s1}\\
	& = & \lim_{\delta \rightarrow 0} \int \frac{1}{\delta}(f(g(\thetav) + (\alpha+\delta) h(\thetav)) - f(g(\thetav) + \alpha h(\thetav))) \diff Q(\thetav)
	\label{EqLimHatrDCT_s2}\\
	& = & \int \lim_{\delta \rightarrow 0} \frac{1}{\delta}(f(g(\thetav) + (\alpha+\delta) h(\thetav)) - f(g(\thetav) + \alpha h(\thetav))) \diff Q(\thetav)
	\label{EqLimHatrDCT_s3}\\
	& = & \int \dot{f}(g(\thetav) + (\alpha+\delta) h(\thetav))  \diff Q(\thetav)
	\label{EqLimHatrDCT_s4}\\
	& < &  \infty,
	\label{EqLimHatrDCT_s5}
\end{IEEEeqnarray}
where the equalities in~\eqref{EqLimHatrDCT_s3} and~\eqref{EqLimHatrDCT_s5} follow from the dominated convergence theorem, Theorem 1.6.9 by~\cite{ash2000probability}. From~\eqref{EqLimHatrDCT_s5}, it follows that the function $\hat{r}$ in~\eqref{EqHatR_lemmAppx} is differentiable at zero. This completes the proof.
\end{IEEEproof}

\begin{lemma}
\label{lemm_f_invIsInc}
Given a strictly convex and differentiable function $f:\set{I}\rightarrow \reals$, the inverse of the derivate of $f$ denoted by the function $\dot{f}^{-1}:\set{J} \rightarrow \set{I}$ is strictly increasing.
\end{lemma}
\begin{IEEEproof}
From the assumption that the function $f:\set{I}\rightarrow \reals$ is strictly convex, it follows from the strict convexity definition that the derivative $\dot{f}: \set{I} \rightarrow \set{J}$ is strictly increasing. Using the continuous inverse theorem by~\cite{bartle2000introduction} in Theorem 5.6, implies that the function $\dot{f}^{-1}:\set{J} \rightarrow \set{I}$ is strictly increasing, which completes the proof.
\end{IEEEproof}

\begin{lemma}
\label{lemm_f_NormInv}
Given a strictly convex and twice differentiable function $f:\set{I}\rightarrow \reals$, and a differentiable function $h:\set{I}^{\star} \rightarrow \set{I}$, for all $x \in \set{I}^{\star}$ it holds that
\begin{IEEEeqnarray}{rCl}
 \frac{\diff }{\diff x}\dot{f}^{-1}(h(x))& = & \frac{\dot{h}(x)}{\ddot{f}(\dot{f}^{-1}(x))}.
\end{IEEEeqnarray}

\end{lemma}
\begin{IEEEproof}
Let the function $g:\set{I}\rightarrow \reals$ be defined for all $x \in \set{I}^{\star}$ by 
\begin{IEEEeqnarray}{rCl}
\label{EqDefgasInv}
g(x) & = & \dot{f}^{-1}(h(x))
\end{IEEEeqnarray}
By the definition of the inverse function, it follows that
\begin{IEEEeqnarray}{rCl}
\label{EqInvofInvIsX}
\dot{f}(g(x))& = & h(x).	
\end{IEEEeqnarray}
Differentiating~\eqref{EqInvofInvIsX} with respect to $x$ yields
\begin{IEEEeqnarray}{rCl}
\label{EqDiffInvofInvIs1}
\frac{\diff }{\diff x}\dot{f}(g(x))
& = & \ddot{f}(g(x))\,\dot{g}(x)\\
& = & \dot{h}(x).	
\end{IEEEeqnarray}
From~\eqref{EqDiffInvofInvIs1} the derivative of the function $g$ in~\eqref{EqInvofInvIsX} is given by
\begin{IEEEeqnarray}{rCl}
\dot{g}(x)
& = & \frac{\dot{h}(x)}{\ddot{f}(g(x))}\label{EqDiffg} \\
& = & \frac{\dot{h}(x)}{\ddot{f}(\dot{f}^{-1}(h(x)))},\label{EqDiffg_s2}
\end{IEEEeqnarray}
where~\eqref{EqDiffg_s2} follows from~\eqref{EqDefgasInv}. Hence, from~\eqref{EqDefgasInv} and~\eqref{EqDiffg_s2} it follows from
\begin{IEEEeqnarray}{rCl}
\frac{\diff }{\diff x}\dot{f}^{-1}(h(x))
& = & \frac{\dot{h}(x)}{\ddot{f}(\dot{f}^{-1}(h(x)))}.
\end{IEEEeqnarray}
This completes the proof.
\end{IEEEproof}

\begin{lemma}
\label{lemm_f_NormInv}
\label{lemm_f_LFTequality}
	The \emph{Legendre-Fenchel} transform of strictly convex and differentiable function $f$, satisfies for all $t \in \set{J}$, with $\set{J}$ in~\eqref{EqDefJinLFT}, 
\begin{IEEEeqnarray}{rCl}
	f^{*}(t) & = &  t\dot{f^{*}}(t)- f\big(\dot{f^{*}}(t)\big).
\end{IEEEeqnarray}
\end{lemma}
\begin{IEEEproof}
From the \emph{Legendre-Fenchel} transform in Definition~\ref{DefLT_cnvxcnj} it holds that for all $t \in \set{J}$, with $\set{J}$ in~\eqref{EqDefJinLFT},
\begin{IEEEeqnarray}{rCl}
\label{EqpfDefLFT}
   f^*(t) & = & \sup_{x \in \set{I}} \left( t z  - f(x) \right).
\end{IEEEeqnarray}
For any $ z \in \set{I}$, setting $ x = z$ yields
\begin{IEEEeqnarray}{rCl}
   f^*(y) & \geq &  y x - f(x),
\end{IEEEeqnarray}   
which rearranges to the Fenchel inequality,
\begin{IEEEeqnarray}{rCl}
\label{EqpfFenchelIneq}
   f(x) + f^{*}(y)  & \geq &  x y,
\end{IEEEeqnarray}
where equality in~\eqref{EqpfFenchelIneq} holds if and only if,
\begin{IEEEeqnarray}{rCl}
\label{EqpfEqualityCondition}
   f^{*}(y) & = &  yx - f(x).
\end{IEEEeqnarray}
By definition of $f^{*}$ in~\eqref{EqpfDefLFT}, the equality $\eqref{EqpfEqualityCondition}$ implies that $x$ achieves the supremum in $f^{*}(y)$, which will be denoted by $x_y$. In other words, $x_y$ is the maximizing argument
\begin{IEEEeqnarray}{rCl}
\label{EqpfMaxArgLFT}
   x_y & = & \arg \max_{x \in \set{I}} xy-f(x).
\end{IEEEeqnarray}
Note that under Assumption~\ref{assum:a}, the solution to the maximization problem
\begin{IEEEeqnarray}{rCl}
\max_{x \in \set{I}} & \ & xy-f(x),
\end{IEEEeqnarray}
is unique, and satisfies
\begin{IEEEeqnarray}{rCl}
\frac{\diff}{\diff x} (xy-f(x)) & = & y - \dot{f}(x) \\
& = & 0. \label{EqpfZeroLFT_Dfproof}
\end{IEEEeqnarray}
From~\eqref{EqpfZeroLFT_Dfproof}, the maximizing argument $x_y$ in~\eqref{EqpfMaxArgLFT} satisfies
\begin{IEEEeqnarray}{rCl}
x_y & = & \dot{f}^{-1}(y)\\
	& = & \dot{f^{*}}(y),
	\label{EqpfSolutionDfDt}
\end{IEEEeqnarray} 
where~\eqref{EqpfSolutionDfDt} follows from \cite[Corollary 23.5.1]{rockafellar1970conjugate}.
Hence, from~\eqref{EqpfSolutionDfDt}, the \emph{Legendre-Fenchel} transform of function the $f$, under Assumption~\ref{assum:a}, satisfies for all $t \in \set{J}$, 
%
\begin{IEEEeqnarray}{rCl}
	f^{*}(t) & = &  t\dot{f^{*}}(t)- f\big(\dot{f^{*}}(t)\big),
\end{IEEEeqnarray}
which completes the proof.
\end{IEEEproof}

\begin{theorem}
\label{Theo_ERM_fDR_LT}
The probability measure $\Pgibbs{P}{Q}$in~\eqref{EqGenpdffDv} satisfies
\begin{IEEEeqnarray}{rCl}
\IEEEeqnarraymulticol{3}{l}{
	\foo{R}_{\dset{z}}(\Pgibbs{P}{Q}) +\lambda \Divf{\Pgibbs{P}{Q}}{Q}
	}\nonumber \\
	& = & -\lambda \int f^{*}(-\frac{\foo{L}_{\dset{z}}(\thetav) + N_{Q,\dset{z}}(\lambda)}{\lambda})\diff Q(\thetav) - N_{Q,\dset{z}}(\lambda),\quad \label{EqPf_ERM_fDR_LT}
\end{IEEEeqnarray}
and 
\begin{IEEEeqnarray}{rCl}
\IEEEeqnarraymulticol{3}{l}{
	\foo{R}_{\dset{z}}(Q) +\lambda\int f(\frac{\diff \Pgibbs{P}{Q}}{\diff Q}(\thetav))\frac{\diff Q}{\diff \Pgibbs{P}{Q}}(\thetav) \diff Q(\thetav)
	}\nonumber \\
	& = & -\lambda\int f^{*}(-\frac{\foo{L}_{\dset{z}}(\thetav) + N_{Q,\dset{z}}(\lambda)}{\lambda})\frac{\diff Q}{\diff \Pgibbs{P}{Q}}(\thetav) \diff Q(\thetav)
	\nonumber \\ &  & -N_{Q,\dset{z}}(\lambda), \quad\label{EqTheo_ERM_fDR_LT_sub2}
\end{IEEEeqnarray}
where $f^{*}$ is the Legendre-Fenchel transform of $f$ (see  Definition~\ref{DefLT_cnvxcnj}), the function $N_{Q,\dset{z}}$ is defined in~\eqref{EqDefNormFunction}, and the functional $\foo{R}_{\dset{z}}$ is defined in~\eqref{EqRxy}.
\end{theorem}
\begin{IEEEproof}
\label{app_theo_ERM_fDR_LT}
The Legendre-Fenchel transform of a strictly convex function $f:\set{I}\rightarrow \reals$ satisfies
\begin{IEEEeqnarray}{rCl}
	f^{*}(t) & \triangleq & \sup_{s\in \set{I}}( ts- f(s)).
\end{IEEEeqnarray}
From Theorem~$23.5$ in~\cite{rockafellar1970conjugate} if $f$ is strictly convex then maximizing argument of the convex conjugate $f^{*}$ satisfies
\begin{IEEEeqnarray}{rCl}
\label{Eq_convcnj_pf}
	f^{*}(t) & = &  t\frac{\diff }{\diff t}f^{*}(t)- f(\frac{\diff }{\diff t}f^{*}(t)).
\end{IEEEeqnarray}
Furthermore, from Corollary~$23.5.1$ in~\cite{rockafellar1970conjugate} the function $\dot{f}^{-1}$ is the derivative of the convex conjugate of~$f$ in~\eqref{Eq_convcnj_pf}, which implies that
Differential Equations
\begin{IEEEeqnarray}{rCl}
	f^{*}(t) & = & t\dot{f}^{-1}(t)- f(\dot{f}^{-1}(t)).
\end{IEEEeqnarray}
From Theorem~\ref{Theo_f_ERMRadNik}, let $t = -\frac{\foo{L}_{\dset{z}}(\thetav)+\beta}{\lambda}$ in the optimization problems in~\eqref{EqOp_f_ERMRERNormal} and~\eqref{EqOp_f_ERM_RND2}, then it holds that for all $\thetav \in \supp Q$,
\begin{IEEEeqnarray}{rCl}
\IEEEeqnarraymulticol{3}{l}{
	f^{*}(-\frac{\foo{L}_{\dset{z}}(\thetav)+\beta}{\lambda}) }
    \nonumber \\
	& = &  -\frac{\foo{L}_{\dset{z}}(\thetav)+\beta}{\lambda}\frac{\diff \Pgibbs{P}{Q}}{\diff Q}(\thetav)- f(\frac{\diff \Pgibbs{P}{Q}}{\diff Q}(\thetav)).\label{Eq_conjugate_between_pf}
\end{IEEEeqnarray}
Taking the integral of~\eqref{Eq_conjugate_between_pf} with respect to the reference measure $Q$, yields
\begin{IEEEeqnarray}{rCl}
\IEEEeqnarraymulticol{3}{l}{
	\int f^{*}(-\frac{\foo{L}_{\dset{z}}(\thetav)+\beta}{\lambda}) \diff Q(\thetav)
    } \nonumber \\
	& = &  \int-\frac{\foo{L}_{\dset{z}}(\thetav)+\beta}{\lambda}\frac{\diff \Pgibbs{P}{Q}}{\diff Q}(\thetav)\diff Q(\thetav) 
   \nonumber \\ & &
    - \int f(\frac{\diff \Pgibbs{P}{Q}}{\diff Q}(\thetav))\diff Q(\thetav)\label{Eq_conjugate_between_pf_s1}\\
	& = & -\frac{1}{\lambda}(\foo{R}_{\dset{z}}(\Pgibbs{P}{Q})+\beta)-\Divf{\Pgibbs{P}{Q}}{Q}\label{Eq_conjugate_between_pf_s2}.
\end{IEEEeqnarray}
Arranging~\eqref{Eq_conjugate_between_pf_s2} results in
\begin{IEEEeqnarray}{rCl}
\IEEEeqnarraymulticol{3}{l}{
	\foo{R}_{\dset{z}}(\Pgibbs{P}{Q}) + \lambda\Divf{\Pgibbs{P}{Q}}{Q}
	}\nonumber \\ 
    & = & -\lambda\int f^{*}(-\frac{\foo{L}_{\dset{z}}(\thetav)+\beta}{\lambda}) \diff Q(\thetav)-\beta ,
\end{IEEEeqnarray}
which completes the proof.

The second part of the proof is as follows.
From Corollary~\ref{coro_mutuallyAbsCont}, equality~\eqref{Eq_conjugate_between_pf} can be rewritten as
\begin{IEEEeqnarray}{rCl}
\frac{\foo{L}_{\dset{z}}(\thetav)+\beta}{\lambda}& = & -f^{*}(-\frac{\foo{L}_{\dset{z}}(\thetav)+\beta}{\lambda})\frac{\diff Q}{\diff \Pgibbs{P}{Q}}(\thetav) 
\nonumber \\ &  & 
- \frac{\diff Q}{\diff \Pgibbs{P}{Q}}(\thetav)f(\frac{\diff \Pgibbs{P}{Q}}{\diff Q}(\thetav)).\label{Eq_conjugate_between_pf2}
\end{IEEEeqnarray}
Taking the integral of~\eqref{Eq_conjugate_between_pf2} with respect to the reference measure $Q$, yields
\begin{IEEEeqnarray}{rCl}
\IEEEeqnarraymulticol{3}{l}{
	\frac{1}{\lambda}\foo{R}_{\dset{z}}(Q) + \frac{\beta}{\lambda} 
    } \nonumber \\
	& = & - \int f^{*}(-\frac{\foo{L}_{\dset{z}}(\thetav)+\beta}{\lambda})\frac{\diff Q}{\diff \Pgibbs{P}{Q}}(\thetav) \diff Q(\thetav)
    \nonumber \\ & &
    - \int \frac{\diff Q}{\diff \Pgibbs{P}{Q}}(\thetav)f(\frac{\diff \Pgibbs{P}{Q}}{\diff Q}(\thetav))\diff Q(\thetav).\label{Eq_conjugate_deff_pf2}
\end{IEEEeqnarray}
Arranging~\eqref{Eq_conjugate_deff_pf2} results in
\begin{IEEEeqnarray}{rCl}
\IEEEeqnarraymulticol{3}{l}{
	\foo{R}_{\dset{z}}(Q) +\lambda\int \frac{\diff Q}{\diff \Pgibbs{P}{Q}}(\thetav)f(\frac{\diff \Pgibbs{P}{Q}}{\diff Q}(\thetav)) \diff Q(\thetav)
	}
    \nonumber \\ & = & -\lambda\int\!\! f^{*}(-\frac{\foo{L}_{\dset{z}}(\thetav)+\beta}{\lambda}) \frac{\diff Q}{\diff \Pgibbs{P}{Q}}(\thetav) \diff Q(\thetav)-\beta ,
\end{IEEEeqnarray}
which completes the proof.
\end{IEEEproof}

In Theorem~\ref{Theo_f_ERMRadNik}, the condition that $\beta$ in~\eqref{EqEqualToABigOne} satisfies \ref{assum:b} in~\eqref{EqDefSetB}, leads to observing that for all $\vect{\theta} \in \supp Q$, 
\begin{IEEEeqnarray}{rCl}
\label{EqJune24at8h26in2024}
\frac{\diff \Pgibbs{P}{Q}}{\diff Q} ( \thetav ) & > & 0, 
\end{IEEEeqnarray}
which leads to the following corollary.%
\begin{corollary}
\label{coro_mutuallyAbsCont}
Under Assumptions \ref{assum:a} and \ref{assum:b}, the probability measures~$Q$ and~$\Pgibbs{P}{Q}$ in~\eqref{EqGenpdffDv} are mutually absolutely continuous.
\end{corollary}

\section{Proof of Theorem~\ref{Theo_f_ERMRadNik}}
%

\begin{IEEEproof}
\label{app_theo_f_ERMRadNik2}
The optimization problem in~\eqref{EqOp_f_ERM_RND2} can be re-written in terms of the \RadonNikodym derivative of the optimization measure $P$ with respect to the reference measure $Q$, denoted by $\frac{\diff P}{\diff Q}: \set{M} \rightarrow [0,\infty)$, which yields:
\begin{subequations}
\label{EqOp_f_ERM_RND2_pf}
\begin{IEEEeqnarray}{cCl}
	\min_{P \in \bigtriangleup_{Q}(\set{M})}
	& \quad & \int \foo{L}_{\dset{z}}(\thetav) \frac{\diff P}{\diff Q}(\thetav) \diff Q(\thetav)\\
	\text{s.t.} 
 	& &  \int f( \frac{\diff P}{\diff Q}(\thetav)) \diff Q(\thetav) \leq \eta \label{EqOp_f_ERM_RND2_pf_c_s1}\\
 	& & \int \frac{\diff P}{\diff Q}(\thetav) \diff Q (\thetav) =1.\label{EqOp_f_ERM_RND2_pf_c_s2}
\end{IEEEeqnarray}
\end{subequations}
The remainder of the proof focuses on the problem in which the optimization is over the Radon-Nikodym derivative $\frac{\diff P}{\diff Q}$ instead of the measures $P$. This is due to the fact that for all $P \in \bigtriangleup_{Q}(\set{M})$, the Radon-Nikodym derivative $\frac{\diff P}{\diff Q}$  is unique up to sets of measure zero with respect to $Q$.
%
%
The first part is as follows. Let $\field{M}$ be the set of measurable functions $\set{M}\rightarrow \reals$ with respect to the measurable space $\msblspc{\set{M}}$ and $(\reals, \borelsigma)$. Let $\field{S}$ be the subset of $\field{M}$, including all nonnegative functions that are absolutely integrable with respect to $Q$. That is, for all $\hat{g}\in \field{S}$, it holds that
\begin{IEEEeqnarray}{rCl}
	\int \abs{\hat{g}(\thetav)}\diff Q(\thetav) & < & \infty.	
\end{IEEEeqnarray}
Note that the set $\field{M}$ forms a real vector space and the set $\field{S}$ is a convex subset of $\field{M}$. Note also that the constraints~\eqref{EqOp_f_ERM_RND2_pf_c_s1} and~\eqref{EqOp_f_ERM_RND2_pf_c_s2} are satisfied by the probability measure $Q$, which also satisfies $Q \in \bigtriangleup_{Q}( \set{M} )$.
Hence, the constraints do not induce an empty feasible set. Finally, note that without loss of generality the minimization in~\eqref{EqOp_f_ERM_RND2_pf} can be written as a minimization problem of the form:
\begin{subequations}
\label{EqOp_f_ERM_Min_all}
\begin{IEEEeqnarray}{rCl}
	\min_{g \in \field{S}} 
	& \quad & \int \foo{L}_{\dset{z}}(\thetav) g(\thetav) \diff Q(\thetav)\\
	\text{s.t.} 
	& &  \frac{1}{\eta}\int f(g(\thetav)) \diff Q(\thetav) \leq 1 \label{EqOp_f_ERM_Min_c_s1}\\
 	& & \int g (\thetav) \diff Q(\thetav) = 1 \label{EqOp_f_ERM_Min_c_s2},
\end{IEEEeqnarray}
\end{subequations}
where the expressions $ \int \foo{L}_{\dset{z}}(\thetav) g(\thetav) \diff Q(\thetav)$ and $\int g (\thetav) \diff Q(\thetav)$ are linear with $g$; the expression $ \frac{1}{\eta}\int f(g(\thetav)) \diff Q(\thetav)$ is convex with $g$.

The proof continues by assuming that the problem in~\eqref{EqOp_f_ERM_Min_all} possesses a solution, which is denoted by $g^{\star} \in \field{S}$. Let $\mu_0 \in [0,\infty)$ be 
\begin{subequations}
\label{EqOp_f_ERM_Min_all_mu}
\begin{IEEEeqnarray}{rcCl}
	\mu_0 &\ \triangleq\ & \min_{g \in \field{S}} 
	 & \int \foo{L}_{\dset{z}}(\thetav) g(\thetav) \diff Q(\thetav)\\
	&   &\text{s.t.}&
	\frac{1}{\eta}\int f(g(\thetav)) \diff Q(\thetav) \leq 1 \label{EqOp_f_ERM_Min_mu_c_s1}\\
 	&   &  & \int g (\thetav) \diff Q(\thetav) = 1 \label{EqOp_f_ERM_Min_mu_c_s2}\\
 	& = &  & \int \foo{L}_{\dset{z}}(\thetav) g^{\star}(\thetav) \diff Q(\thetav).
\end{IEEEeqnarray}
\end{subequations}
From Theorem~1, Section~8.3 in~\cite{luenberger1997bookOptimization}, it holds that there exists two tuples $(a_1,b_1)$ and $(a_2,b_2)$ in $\reals^2$ such that
\begin{subequations}
\label{EqOp_f_ERM_Min_all_anci}
\begin{IEEEeqnarray}{rCl}
	\mu_0 
	& = & \min_{g \in \field{S}} \{ \int \foo{L}_{\dset{z}}(\thetav) g(\thetav) \diff Q(\thetav) +  \frac{a_1}{\eta}\int f(g(\thetav)) \diff Q(\thetav) + b_1 + a2\int g (\thetav) \diff Q(\thetav) + b_2  \},
\end{IEEEeqnarray}
and moreover,
\begin{IEEEeqnarray}{rCl}
\label{EqOp_f_ERM_Min_all_anci_c_s2}
	0 & = & \frac{a_1}{\eta}\int f(g^{\star}(\thetav)) \diff Q(\thetav) + b_1, \text{ and}\\
\label{EqOp_f_ERM_Min_all_anci_c_s3}
	0 & = & a2\int g^{\star} (\thetav) \diff Q(\thetav) + b_2. 
\end{IEEEeqnarray}
\end{subequations}
Hence, the proof continues by solving the ancillary optimization problem in~\eqref{EqOp_f_ERM_Min_all_anci}, which allows the reformulation of the optimization problem in an unconstrained dual problem. This reformulation is possible as the tuples $(a_1,b_1)$ and $(a_2,b_2)$ are such that equalities~\eqref{EqOp_f_ERM_Min_all_anci_c_s2} and~\eqref{EqOp_f_ERM_Min_all_anci_c_s3} are satisfied, by definition.

Let the function $L:\field{S} \rightarrow \reals$ be such that
\begin{IEEEeqnarray}{rCl}
\label{Eq_Lagrange_Min}
	L(g) & = & \int \foo{L}_{\dset{z}}(\thetav) g(\thetav) \diff Q(\thetav) +  \frac{a_1}{\eta}\int f(g(\thetav))\diff Q(\thetav) + b_1   + a2\int g (\thetav) \diff Q(\thetav) + b_2. 	
\end{IEEEeqnarray}
Let $\hat{g}:\set{M} \rightarrow \reals$ be a function in $\field{S}$. The G{\^a}teaux differential of the functional $L$ in~\eqref{Eq_Lagrange_Min} at $\left(g, \beta\right) \in \mathscr{M}\times \reals$ in the direction of $\hat{g}$, if it exists, is
\begin{IEEEeqnarray}{rcl}
\label{EqNecessaryCondtionDivff}
\partial L(g; \hat{g} ) & \triangleq & \left.\frac{\diff}{\diff \alpha}  L(g + \alpha \hat{g}, \beta) \right|_{\alpha = 0}.
\end{IEEEeqnarray}
The proof continues under the assumption that the function $g$ and $\hat{g}$ are such that the G{\^a}teaux differential in~\eqref{EqNecessaryCondtionDivff} exists. Under such an assumption, let the function $r: \reals \rightarrow \reals$ satisfy for all $\alpha \in (-\epsilon, \epsilon)$, with $\epsilon$ arbitrarily small, that
\begin{IEEEeqnarray}{rcl}
r(\alpha)
& = & \int \foo{L}_{\vect{z}}(\thetav)(g (\thetav) + \alpha \hat{g}(\thetav))\,\diff Q(\thetav) 
+ \> \frac{a_1}{\eta} \int f(g(\thetav) + \alpha \hat{g}(\thetav))\,\diff Q(\thetav)+b_1 
+ \> a_2\int(g(\thetav)  + \alpha \hat{g}(\thetav))\diff Q(\thetav) + b_2 
\end{IEEEeqnarray}
which can be rewritten as follows,
\begin{IEEEeqnarray}{rcl}
r(\alpha) 
& = & \alpha \int \hat{g}(\thetav)(a_2 +\foo{L}_{\vect{z}}(\thetav))   \diff Q(\thetav)
+ \> \frac{a_1}{\eta} \int f(g(\thetav) + \alpha \hat{g}(\thetav))\diff Q(\thetav) 
+ \int g(\thetav)(a_2 + \foo{L}_{\vect{z}}(\thetav))\,\diff Q(\thetav)
+ b_1  + b_2.\label{EqpreGateaux_r}
\end{IEEEeqnarray}
Note that the first terms in~\eqref{EqpreGateaux_r} is linear with $\alpha$; the second term can be written using the function $\hat{r}: \reals \to \reals$ in~\eqref{EqHatR_lemmAppx} such that for all $\alpha \in (-\epsilon, \epsilon)$, with $\epsilon$ arbitrarily small, it holds that
\begin{IEEEeqnarray}{rcl}
\label{EqrHatForDaunas}
\hat{r}(\alpha) & = &  \lambda \int f(g(\thetav) + \alpha \hat{g}(\thetav))\,\diff Q(\thetav);
\end{IEEEeqnarray}
and the remaining terms are independent of $\alpha$.

Hence, based on the fact that the function $\hat{r}$ in~\eqref{EqrHatForDaunas} is differentiable at zero (see Lemma~\ref{lemm_ERM_fDR_DiffZero}), so is the function $r$ in~\eqref{EqpreGateaux_r}, which implies that the G{\^a}teaux differential of $\partial L (g,\hat{g})$ in~\eqref{EqNecessaryCondtionDivff} exists.
The derivative of the real function $r$ in~\eqref{EqpreGateaux_r} is
\begin{IEEEeqnarray}{rCl}
	\frac{\diff}{\diff \alpha} r(\alpha) 
	& = & \frac{\diff}{\diff \alpha} \left(\alpha \int \hat{g}(\thetav)(a_2 +\foo{L}_{\vect{z}}(\thetav))   \diff Q(\thetav)
	+ \> \frac{a_1}{\eta} \int f(g(\thetav) + \alpha \hat{g}(\thetav))\diff Q(\thetav) 
	\right. \nonumber \\ &    & \left. 
	+ \int g(\thetav)(a_2 + \foo{L}_{\vect{z}}(\thetav) \right)\,\diff Q(\thetav)
	+ b_1  + b_2)
	\label{Eq_r_dalpha_Min_s1}\\
	& = &  \int \hat{g}(\thetav)(a_2 +\foo{L}_{\vect{z}}(\thetav))   \diff Q(\thetav) 
	+  \frac{a_1}{\eta} \int \frac{\diff}{\diff \alpha} f(g(\thetav) + \alpha \hat{g}(\thetav))\diff Q(\thetav)
	\label{Eq_r_dalpha_Min_s2}\\
	& = &  \int \hat{g}(\thetav)(a_2 +\foo{L}_{\vect{z}}(\thetav))   \diff Q(\thetav) 
	+  \frac{a_1}{\eta} \int \hat{g}(\thetav)\dot{f}(g(\thetav) + \alpha \hat{g}(\thetav))\diff Q(\thetav)
	\label{Eq_r_dalpha_Min_s3}
\end{IEEEeqnarray}
where~\eqref{Eq_r_dalpha_Min_s2} follows from Theorem~\ref{Theo_ERM_fDR_Leibniz}.
From equations~\eqref{EqNecessaryCondtionDivff} and~\eqref{Eq_r_dalpha_Min_s3}, it follows that
\begin{subequations}
\label{Eq_dL_Min}
\begin{IEEEeqnarray}{rCl}
	\partial L(g;h) 
	& = &  \int \hat{g}(\thetav)(a_2 +\foo{L}_{\vect{z}}(\thetav))   \diff Q(\thetav) 
	+  \frac{a_1}{\eta} \int \hat{g}(\thetav)\dot{f}(g(\thetav))\diff Q(\thetav)
	\label{Eq_dL_Min_s1}\\
	& = &  \int \hat{g}(\thetav)(a_2 +\foo{L}_{\vect{z}}(\thetav) 
	+  \frac{a_1}{\eta}  \dot{f}(g(\thetav)))\diff Q(\thetav)
	\label{Eq_dL_Min_s2}.
\end{IEEEeqnarray}
\end{subequations}
A necessary condition to use Theorem~1 Chapter 7 in~\cite{luenberger1997bookOptimization} for the functional $L$ in~\eqref{Eq_Lagrange_Min} to have a minimum at $g^{\star}$ is that for all functions $\hat{g}\in \field{S}$,
\begin{IEEEeqnarray}{rCl}
\label{Eq_dL_Min_luen_Th1}
\partial L(g^{\star};h) & = & 0.	
\end{IEEEeqnarray}
From~\eqref{Eq_dL_Min_luen_Th1}, it follows that $\Pgibbs{}{}$ must satisfy for all functions $\hat{g}$ in $\field{S}$ that
\begin{IEEEeqnarray}{rCl}
	\label{Eq_dL_Min_zero}
	\foo{L}_{\dset{z}}(\thetav)  + \frac{a_1}{\eta}\dot{f}(g(\thetav)) + a_2 & = & 0.	
\end{IEEEeqnarray}
Assuming that
\begin{IEEEeqnarray}{rCl}
\label{Eq_dL_Min_zero_c}
	a_1 \neq 0,	
\end{IEEEeqnarray}
From~\eqref{Eq_dL_Min_zero}, it follows that
\begin{IEEEeqnarray}{rCl}
	\label{Eq_Op_RND_Min}
	g^{\star}(\thetav) & = & \dot{f}^{-1}(-\frac{\eta}{a_1}(\foo{L}_{\dset{z}}(\thetav) + a_2)),	
\end{IEEEeqnarray}
where the values $a_1$ and $a_2$ satisfy~\eqref{EqOp_f_ERM_Min_all_anci_c_s2} and~\eqref{EqOp_f_ERM_Min_all_anci_c_s3} and~\eqref{Eq_dL_Min_zero_c}.

The remainder of the proof focuses on determining the values of $a_1$, $a_2$, $b_1$, and $b_2$, which must also be such that $g^{\star}$ in~\eqref{Eq_Op_RND_Min} satisfies the constraints~\eqref{EqOp_f_ERM_Min_c_s1} and~\eqref{EqOp_f_ERM_Min_c_s2} under the assumption that $\foo{L}_{\dset{z}}$ in~\eqref{EqLxy} is separable. 
For instance, from constraints~\eqref{EqOp_f_ERM_Min_c_s1} and~\eqref{EqOp_f_ERM_Min_all_anci_c_s3} it follows that
\begin{IEEEeqnarray}{rCl}
\label{EqConstrainA2eqB2}
	a_2 & = & -b_2.
\end{IEEEeqnarray}
From~\eqref{EqConstrainA2eqB2}, the constraint in~\eqref{EqOp_f_ERM_Min_all_anci_c_s3} implies that the choice of $a_2$ satisfies
\begin{IEEEeqnarray}{rCl}
\label{EqConstA2isOne}
	1 & = &	\int g^{\star}(\thetav)\diff Q(\thetav).
\end{IEEEeqnarray}
Similarly, the function $g^{\star}$ in~\eqref{Eq_Op_RND_Min} is the Radon-Nikodym derivative with respect to $Q$ of the solution $P^{\star} \in \bigtriangleup( \set{M} )$ to the problem in~\eqref{EqOp_f_ERM_RND2_pf}. Hence,~\eqref{EqOp_f_ERM_Min_all_anci_c_s2} can be written as follows
\begin{IEEEeqnarray}{rCl}
\label{Eq_Op_KL_Min_Jpz}
	\frac{a_1}{\eta}\KL{P^{\star}}{Q}  + b_1 & = & 0,	
\end{IEEEeqnarray}
which implies
\begin{IEEEeqnarray}{rCl}
	b_1 & = & - \frac{a_1}{\eta} \KL{P^{\star}}{Q}.	
\end{IEEEeqnarray}
Considering $a_1$, note that if $a_1 < 0$, given two models $\thetav_1$ and $\thetav_2$ in $\set{M}$, such that $\foo{L}_{\dset{z}}(\thetav_1) \leq \foo{L}_{\dset{z}}(\thetav_2)$, it holds that
\begin{IEEEeqnarray}{rCl}
\label{EqIneq_NotIncreasingInput}
	-\frac{a_1}{\eta}(\foo{L}_{\dset{z}}(\thetav_1) +a_2)
	& < & -\frac{a_1}{\eta}(\foo{L}_{\dset{z}}(\thetav_2) +a_2).	
\end{IEEEeqnarray}
From Lemma~\ref{lemm_f_invIsInc},~\eqref{Eq_Op_RND_Min} implies that the function $g^{\star}$ is strictly increasing.
Hence, under the assumption that $a_1 < 0$, inequality~\eqref{EqIneq_NotIncreasingInput} implies that
\begin{IEEEeqnarray}{rCl}
\label{EqIneq_g_NotIncreasing}
	g^{\star}(\thetav_1) & < & g^{\star}(\thetav_2).	
\end{IEEEeqnarray}
Since $g^{\star}$ is strictly increasing and positive, by observing that the expected empirical risk is a weighted average in which models $\thetav \in \supp Q$ that induce larger values of empirical risk $\foo{L}_{\dset{z}}$ are weighted more heavily implies that
\begin{IEEEeqnarray}{rCl}
\label{Eqa1Contradicts}
	\int \foo{L}_{\dset{z}}(\thetav)\diff Q(\thetav) & < & \int \foo{L}_{\dset{z}}(\thetav) g^{\star}(\thetav)\diff Q(\thetav),	
\end{IEEEeqnarray}
which is a contradiction.
Thus, the focus in the remainder of the proof is the case in which
\begin{IEEEeqnarray}{rCl}
	a_1 & > & 0,	
\end{IEEEeqnarray}
which implies that for the models $\thetav_1$ and $\thetav_2$ in $\supp Q$ such that $\foo{L}_{\dset{z}}(\thetav_1) \leq \foo{L}_{\dset{z}}(\thetav_2)$, it holds that
\begin{IEEEeqnarray}{rCl}
\label{EqIneq_g_decreasing}
	g^{\star}(\thetav_1) & \geq & g^{\star}(\thetav_2).	
\end{IEEEeqnarray}
Given the pairs $(a_1,a_2)$ and $(\hat{a}_1,\hat{a}_2)$ in $\reals^2$ such that each pair satisfies the constraints in~\eqref{EqOp_f_ERM_Min_all_anci_c_s2} and~\eqref{EqOp_f_ERM_Min_all_anci_c_s3}, then from~\eqref{Eq_Op_RND_Min} there exist a solution for each pair given by
\begin{IEEEeqnarray}{rCl}
	\label{Eq_Op_RND_Min_sub1}
	g^{\star}(\thetav) & = & \dot{f}^{-1}(-\frac{\eta}{a_1}(\foo{L}_{\dset{z}}(\thetav) + a_2)),	
\end{IEEEeqnarray}
and
\begin{IEEEeqnarray}{rCl}
	\label{Eq_Op_RND_Min_sub2}
	\hat{g}^{\star}(\thetav) & = & \dot{f}^{-1}(-\frac{\eta}{\hat{a}_1}(\foo{L}_{\dset{z}}(\thetav) + \hat{a}_2)),	
\end{IEEEeqnarray}
where the functions $g^{\star}$ and $\hat{g}^{\star}$ are the Radon-Nikodym derivative of the solutions $P^{\star}$ and $\hat{P}^{\star}$ with respect to $Q$ for each pair $(a_1,a_2)$ and $(\hat{a}_1,\hat{a}_2)$, respectively.
Under the assumption that $a_1 < \hat{a}_1$, it holds that for all $\thetav \in \supp Q$,
\begin{IEEEeqnarray}{rCl}
	-\frac{\eta}{a_1}(\foo{L}_{\dset{z}}(\thetav) + a_2) 
	& < & -\frac{\eta}{ \hat{a}_1}(\foo{L}_{\dset{z}}(\thetav) + a_2),
\end{IEEEeqnarray}
which from Lemma~\ref{lemm_f_invIsInc} implies that
\begin{IEEEeqnarray}{rCl}
\label{EqIneq_cnvxcnj}
	\dot{f}^{-1}(-\frac{\eta}{a_1}(\foo{L}_{\dset{z}}(\thetav) + a_2))  & < & \dot{f}^{-1}(-\frac{\eta}{\hat{a}_1}(\foo{L}_{\dset{z}}(\thetav) + a_2)).
\end{IEEEeqnarray}
From~\eqref{EqIneq_cnvxcnj}, it holds that
\begin{IEEEeqnarray}{rCl}
	1 & = & \int  \dot{f}^{-1}(-\frac{\eta}{a_1}(\foo{L}_{\dset{z}}(\thetav) + a_2)) \diff Q(\thetav)\\
	  & < & \int  \dot{f}^{-1}(-\frac{\eta}{\hat{a}_1}(\foo{L}_{\dset{z}}(\thetav) + a_2)) \diff Q(\thetav).
\end{IEEEeqnarray}
Then, for the pair $(\hat{a}_1,\hat{a}_2)$ to satisfy,
\begin{IEEEeqnarray}{rCl}
	\int  \dot{f}^{-1}(-\frac{\eta}{\hat{a}_1}(\foo{L}_{\dset{z}}(\thetav) + \hat{a}_2)) \diff Q(\thetav) & = & 1,
\end{IEEEeqnarray}
under the assumption that $a_1 < \hat{a}_1$, the value $\hat{a}_2$  must satisfy $a_2 < \hat{a}_2$.
Using the fact that $0<a_1<\hat{a}_1$ and $a_2<\hat{a}_2$, consider the partition of the set $\set{M}$ formed by the sets $\set{A}_0$, $\set{A}_1$ and $\set{A}_2$, which satisfy the following:
\begin{IEEEeqnarray}{rCl}
	\set{A}_0 & \triangleq & \{ \thetav \in \set{M}: \foo{L}_{\dset{z}}(\thetav) = \frac{a_2\hat{a}_1-\hat{a}_2a_1}{a_1-\hat{a}_1}\},
	\label{EqproofSetA0}\\
	\set{A}_1 & \triangleq  &\{ \thetav \in \set{M}: \foo{L}_{\dset{z}}(\thetav) < \frac{a_2\hat{a}_1-\hat{a}_2a_1}{a_1-\hat{a}_1}\},
	\label{EqproofSetA1}\\
	\set{A}_2 & \triangleq  &\{ \thetav \in \set{M}: \foo{L}_{\dset{z}}(\thetav) > \frac{a_2\hat{a}_1-\hat{a}_2a_1}{a_1-\hat{a}_1}\}.
	\label{EqproofSetA2}
\end{IEEEeqnarray}
Note that for all $\thetav \in \set{A}_0$, the pair $(\hat{a}_1,\hat{a}_2)$ satisfies
\begin{IEEEeqnarray}{rCl}
	-\frac{\eta}{\hat{a}_1}(\foo{L}_{\dset{z}}(\thetav) + \hat{a}_2) 
	& = & -\frac{\eta}{\hat{a}_1}(\frac{a_2\hat{a}_1-\hat{a}_2a_1}{a_1-\hat{a}_1} + \hat{a}_2)
	\label{EqprehatEqualLxy_s1}\\
	& = & -\frac{\eta}{\hat{a}_1}(\frac{a_2\hat{a}_1-\hat{a}_2a_1}{a_1-\hat{a}_1}+\frac{ \hat{a}_2(a_1-\hat{a}_1)}{a_1-\hat{a}_1})
	\label{EqprehatEqualLxy_s2}\\
	& = & -\frac{\eta}{\hat{a}_1}(\frac{a_2\hat{a}_1-\hat{a}_2 \hat{a}_1}{a_1-\hat{a}_1})
	\label{EqprehatEqualLxy_s3}\\
	& = & -\eta(\frac{a_2-\hat{a}_2}{a_1-\hat{a}_1}).
	\label{EqprehatEqualLxy_s4}
\end{IEEEeqnarray}
Similarly, for all $\thetav \in \set{A}_0$, the pair $(a_1,a_2)$ satisfies
\begin{IEEEeqnarray}{rCl}
	-\frac{\eta}{a_1}(\foo{L}_{\dset{z}}(\thetav) + a_2) 
	& = & -\frac{\eta}{a_1}(\frac{a_2\hat{a}_1-\hat{a}_2a_1}{a_1-\hat{a}_1} + a_2)
	\label{EqpreEqualLxy_s1}\\
	& = & -\frac{\eta}{a_1}(\frac{a_2\hat{a}_1-\hat{a}_2a_1}{a_1-\hat{a}_1}+\frac{ a_2(a_1-\hat{a}_1)}{a_1-\hat{a}_1})
	\label{EqpreEqualLxy_s2}\\
	& = & -\frac{\eta}{a_1}(\frac{a_2a_1-\hat{a}_2 a_1}{a_1-\hat{a}_1})
	\label{EqpreEqualLxy_s3}\\
	& = & -\eta(\frac{a_2-\hat{a}_2}{a_1-\hat{a}_1}).
	\label{EqpreEqualLxy_s4}
\end{IEEEeqnarray}
Hence, from~\eqref{EqprehatEqualLxy_s4} and~\eqref{EqpreEqualLxy_s4} for all $\thetav \in \set{A}_0$, it holds that
\begin{IEEEeqnarray}{rCl}
\label{EqEqualandhatl_Lxy}
	-\frac{\eta}{\hat{a}_1}(\foo{L}_{\dset{z}}(\thetav) + \hat{a}_2) 
	& = & -\frac{\eta}{a_1}(\foo{L}_{\dset{z}}(\thetav) + a_2) .
\end{IEEEeqnarray}
Then, from~\eqref{EqEqualandhatl_Lxy} it follows that for all  $\thetav \in \set{A}_0$ it holds that
\begin{subequations}
\label{EqEqualLxy}
\begin{IEEEeqnarray}{rCl}
	g^{\star}(\thetav) & = & \dot{f}^{-1}(-\frac{\eta}{a_1}(\foo{L}_{\dset{z}}(\thetav) + a_2))\label{EqEqualLxy_s1}\\
	& = &\dot{f}^{-1}(-\frac{\eta}{\hat{a}_1}(\foo{L}_{\dset{z}}(\thetav) + \hat{a}_2))\label{EqEqualLxy_s2}\\
	& = & \hat{g}^{\star}(\thetav)\label{EqEqualLxy_s3},
\end{IEEEeqnarray}
\end{subequations}
where~\eqref{EqEqualLxy_s2} follows from the fact that $\dot{f}^{-1}$ in~\eqref{Eq_Op_RND_Min} is strictly increasing.
Therefore, for all~$\thetav \in \set{A}_1$, it holds that
\begin{IEEEeqnarray}{rCl}
\label{EqIneqRDN_f_A1}
	g^{\star}(\thetav) & > & \hat{g}^{\star}(\thetav), 
\end{IEEEeqnarray}
and for all~$\thetav \in \set{A}_2$, it holds that
\begin{IEEEeqnarray}{rCl}
\label{EqIneqRDN_f_A2}
	g^{\star}(\thetav) & < & \hat{g}^{\star}(\thetav), 
\end{IEEEeqnarray}
where inequalities~\eqref{EqIneqRDN_f_A1} and~\eqref{EqIneqRDN_f_A2} follow from~\eqref{EqEqualLxy} and the fact that $\dot{f}^{-1}$ in~\eqref{Eq_Op_RND_Min} is strictly increasing (see Lemma~\ref{lemm_f_invIsInc}).
Let $P^{\star}$ and $\hat{P}^{\star}$ denote the probability measures defined by the pairs $(a_1,a_2)$ and $(\hat{a}_1, \hat{a}_2)$, respectively. From~\eqref{Eq_Op_RND_Min_sub1} and~\eqref{Eq_Op_RND_Min_sub2} it follows that 
\begin{IEEEeqnarray}{rCl}
	P^{\star}(\set{A}_1) 
	& = & \int_{\set{A}_1} g^{\star}(\thetav) \diff Q(\thetav) 
	\label{EqpreIneqP_A1_s1},
\end{IEEEeqnarray}
and
\begin{IEEEeqnarray}{rCl}
	\hat{P}^{\star}(\set{A}_1) 
	& = & \int_{\set{A}_1} \hat{g}^{\star}(\thetav) \diff Q(\thetav) 
	\label{EqpreIneqhatP_A1_s1}.
\end{IEEEeqnarray}
From~\eqref{EqpreIneqP_A1_s1} and~\eqref{EqpreIneqhatP_A1_s1} the measures $P^{\star}$ and $\hat{P}^{\star}$ over the set $\set{A}_1$ in~\eqref{EqproofSetA1} satisfy
\begin{subequations}
\label{EqIneqP_A1}
\begin{IEEEeqnarray}{rCl}
	P^{\star}(\set{A}_1) 
	& = & \int_{\set{A}_1} g^{\star}(\thetav) \diff Q(\thetav) 
	\label{EqIneqP_A1_s1} \\  
	& > & \int_{\set{A}_1} \hat{g}^{\star}(\thetav)  \diff Q(\thetav)
	\label{EqIneqP_A1_s2}\\
	& = & \hat{P}^{\star}(\set{A}_1),
	\label{EqIneqP_A1_s3}
\end{IEEEeqnarray}
\end{subequations}
where~\eqref{EqIneqP_A1_s2} follows from~\eqref{EqIneqRDN_f_A1}.
Similarly, from~\eqref{EqpreIneqP_A1_s1} and~\eqref{EqpreIneqhatP_A1_s1}, the measures $P^{\star}$ and $\hat{P}^{\star}$ over the set $\set{A}_2$ in~\eqref{EqproofSetA2} satisfy
\begin{subequations}
\label{EqIneqP_A2}
\begin{IEEEeqnarray}{rCl}
	P^{\star}(\set{A}_2)
	& = & \int_{\set{A}_2} g^{\star}(\thetav) \diff Q(\thetav) 
	\label{EqIneqP_A2_s1} \\  
	& < & \int_{\set{A}_2} \hat{g}^{\star}(\thetav)  \diff Q(\thetav)
	\label{EqIneqP_A2_s2}\\
	& = & \hat{P}^{\star}(\set{A}_2).
\end{IEEEeqnarray}
\end{subequations}
Hence, from~\eqref{EqIneqP_A1} and~\eqref{EqIneqP_A2} the expected empirical risk of the measures $P^{\star}$ and $\hat{P}^{\star}$ satisfies
\begin{IEEEeqnarray}{rCl}
\label{EqIneqLxy_pf}
	\foo{R}_{\dset{z}}(P^{\star}) & < & \foo{R}_{\dset{z}}(\hat{P}^{\star}). 
\end{IEEEeqnarray}
Observe that from~\eqref{EqIneqLxy_pf} and the assumption of $a_1 < \hat{a}_1$, it follows that 
\begin{subequations}
\label{Eq_da1_g_star_increase}
\begin{IEEEeqnarray}{rCl}
	\frac{\diff }{\diff a_1} \foo{R}_{\dset{z}}(P^{\star}) & = & \frac{\diff }{\diff a_1} \int \foo{L}_{\dset{z}}(\thetav) \diff P^{\star}(\thetav)
	\label{Eq_da1_g_star_increase_s1}\\
	& = & \lim_{\hat{a}_1 \rightarrow a_1} \frac{\int \foo{L}_{\dset{z}}(\thetav) \diff \hat{P}^{\star}(\thetav)-\int \foo{L}_{\dset{z}}(\thetav) \diff P^{\star}(\thetav)}{\hat{a}_1 - a_1}
	\label{Eq_da1_g_star_increase_s2}\\
	& > & 0,\label{Eq_da1_g_star_increase_s3}
\end{IEEEeqnarray}
\end{subequations}
where~\eqref{Eq_da1_g_star_increase_s3} follows from~\eqref{EqIneqLxy_pf}.
From~\eqref{EqIneq_g_decreasing},~\eqref{EqIneqRDN_f_A1} and~\eqref{EqIneqRDN_f_A2}, for all models $(\thetav_1,\thetav_2)\in \set{A}_1\times\set{A}_2$ it follows that
\begin{IEEEeqnarray}{rCl}
\label{EqPreDa_g}
	g^{\star}(\thetav_1)-g^{\star}(\thetav_2)
	& > & \hat{g}^{\star}(\thetav_1)-\hat{g}^{\star}(\thetav_2).
\end{IEEEeqnarray}
Furthermore, from~\eqref{EqPreDa_g} and Lemma~\ref{lemm_f_invIsInc}, for all models $(\thetav_1,\thetav_2)\in (\supp Q)^2$, such that $\foo{L}_{\dset{z}}(\thetav_1) < \foo{L}_{\dset{z}}(\thetav_2)$, it follows that
\begin{IEEEeqnarray}{rCl}
	\label{EqPreDa_g_all}
	g^{\star}(\thetav_1)-g^{\star}(\thetav_2)
	& > & \hat{g}^{\star}(\thetav_1)-\hat{g}^{\star}(\thetav_2)>0.
\end{IEEEeqnarray}
From~\eqref{EqPreDa_g_all} and the assumption that $f$ is strictly convex, it holds that for all $\thetav \in \supp Q$
\begin{IEEEeqnarray}{rCl}
	\label{EqPreDa_Div_f}
	f(g^{\star}(\thetav)) & > & f(\hat{g}^{\star}(\thetav)).
\end{IEEEeqnarray}
Observe that from~\eqref{EqPreDa_Div_f} and the assumption of $a_1 < \hat{a}_1$, it follows that
\begin{IEEEeqnarray}{rCl}
	\frac{\diff }{\diff a_1} f(g^{\star}(\thetav)) 
	& = & \lim_{\hat{a}_1 \rightarrow a_1} \frac{ f(g^{\star}(\thetav))- f(\hat{g}^{\star}(\thetav))}{a_1 - \hat{a}_1 }
	\label{Eq_da1_Div_f_decrease_s1}\\
	& < & 0,\label{Eq_da1_Div_f_decrease_s3}
\end{IEEEeqnarray}
which implies that $f(g^{\star})$ in~\eqref{Eq_Op_RND_Min} is strictly decreasing with respect to $a_1$. 
Note also that
\begin{IEEEeqnarray}{rCl}
	\frac{\diff }{\diff a_1} \Divf{P^{\star}}{Q} 
	& = &\frac{\diff }{\diff a_1} \int f(g^{\star}(\thetav)) \diff Q(\thetav)\label{Eq_Op_da1_Divf_s1}\\
	& = & \int \frac{\diff }{\diff a_1}f(g^{\star}(\thetav)) \diff Q(\thetav)\label{Eq_Op_da1_Divf_s2}\\
	& < & 0,\label{Eq_Op_da1_Divf_s7}
\end{IEEEeqnarray}
where~\eqref{Eq_Op_da1_Divf_s2} follows from the dominated convergence theorem by~\cite{ash2000probability} in Theorem~1.6.9, 
and~\eqref{Eq_Op_da1_Divf_s7} follows from~\eqref{Eq_da1_Div_f_decrease_s3}.
Hence, from~\eqref{Eq_da1_g_star_increase} the terms $\int g^{\star}(\thetav)\foo{L}_{\dset{z}}(\thetav) \diff Q(\thetav)$ in~\eqref{EqOp_f_ERM_RND2_pf_c_s1} is strictly increasing with $a_1$ and from~\eqref{Eq_Op_da1_Divf_s7} the term $\int f(g^{\star}(\thetav)) \diff Q(\thetav)$ in~\eqref{EqOp_f_ERM_RND2_pf_c_s2} is strictly decreasing with $a_1$. This implies that $a_1 > 0$ shall be chosen such that
\begin{IEEEeqnarray}{rCl}
	\KL{P^{\star}_1}{Q} = \eta,
\end{IEEEeqnarray}
and justify the uniqueness of the solution.

For the case in which the empirical risk function $\foo{L}_{\dset{z}}$ in~\eqref{EqLxy} is nonseparable (see Definition \ref{Def_SeparableLxy}), the objective function in~\eqref{EqOp_f_ERM_RND2} is a constant and thus the problem is ill-posted.

Therefore, choosing a real value $\lambda = \frac{a_1}{\eta}$, the real $a_2$ to satisfy 
\begin{equation}
\label{EqSetB_a_2}
a_2 \in \left\lbrace t\in \reals: \forall \vect{\theta} \in \supp Q , 0 <  \dot{f}^{-1} \left( -\frac{t + \foo{L}_{\vect{z}}(\thetav)}{\lambda} \right)\right\rbrace, 
\end{equation}
and denoting the solution $P^{\star}$ as $\Pgibbs{P}{Q}$, it holds that $g^{\star}$ in~\eqref{Eq_Op_RND_Min} can be written as $\frac{\diff \Pgibbs{P}{Q}}{\diff Q}$, and thus, for all $(\thetav) \in \supp Q$,
\begin{IEEEeqnarray}{rCl}
	\frac{\diff \Pgibbs{P}{Q}}{\diff Q} (\thetav) 
	& = & \dot{f}^{-1}(-\frac{\foo{L}_{\dset{z}}(\thetav)+a_2}{\lambda}),
\end{IEEEeqnarray}
where $\lambda$ is such that $\Divf{\Pgibbs{P}{Q}}{Q}=\eta$.
This completes the proof.
\end{IEEEproof}
\section{Proofs of Section~\ref{sec:analysisRegFact}}
\label{sec:AppSec4}
\subsection{Proof of Theorem~\ref{theo_InfDevKfDR}}
\begin{IEEEproof}
\label{AppProofLemmaInfDevKDivf}
The proof is divided into two parts.
The first part uses the properties of the $f$-divergences regularization to prove that the normalization function $N_{Q, \dset{z}}:\set{A}_{Q, \dset{z}} \rightarrow \set{B}_{Q, \dset{z}}$ in~\eqref{EqDefNormFunction} is strictly increasing.
The second part proves the continuity of the function $N_{Q, \dset{z}}$ in~\eqref{EqDivfKrescaling}.

The first part is as follows.
Given a pair $(a,b) \in \set{A}_{Q, \dset{z}}\times\set{B}_{Q, \dset{z}}$, with $\set{A}_{Q, \dset{z}}$ in~\eqref{EqDefNormFunction}, assume that
\begin{equation}
\label{EqProofKrescalingL1}
N_{Q, \dset{z}}(a) = b.
\end{equation}
This implies that
\begin{IEEEeqnarray}{rCl}
1 
& = & \int \frac{\diff \Pgibbs[\dset{z}][b]{P}{Q}}{\diff Q}(\thetav) \diff Q(\thetav)
\label{Eq_ProofLambdaIsTheInvOfKbar_s1}\\
& = & \int \dot{f}^{-1}(-\frac{b + \foo{L}_{\dset{z}}(\thetav)}{a})\diff Q(\thetav).
\label{Eq_ProofLambdaIsTheInvOfKbar_s2}
\end{IEEEeqnarray}
Note that the inverse $\dot{f}^{-1}$ exists from the fact that $f$ is strictly convex, which implies that $\dot{f}$ is a strictly increasing function. Hence, $\dot{f}^{-1}$ is also a strictly increasing function in $\set{B}_{Q, \dset{z}}$ based on Theorem 5.6.9 in~\cite{bartle2000introduction}.
Moreover, from the assumption that $f$ is strictly convex and differentiable, it holds that $\dot{f}$ is continuous, which follows from Proposition 5.44 in~\cite{douchet2010Analyse}. This implies that $\dot{f}^{-1}$ is continuous.
From~Lemma~\ref{lemm_f_invIsInc} the function $\dot{f}^{-1}$ is strictly increasing such that for all $b \in \set{B}_{Q, \dset{z}}$ and for all $\thetav \in \supp Q$, it holds that
\begin{IEEEeqnarray}{rCl}
\label{Eq_ProofFinitenes_pf}	
\dot{f}^{-1}(- \frac{b +\foo{L}_{\dset{z}}(\thetav)}{a})
& \leq  & \dot{f}^{-1}(- \frac{b +\delta^\star_{Q, \dset{z}}}{a}),
\end{IEEEeqnarray}
with $\delta^\star_{Q, \dset{z}}$ defined in~\eqref{EqDefDeltaStar}.
Then, from~\eqref{Eq_ProofFinitenes_pf} it follows that
\begin{IEEEeqnarray}{rCl}
\int \dot{f}^{-1}(- \frac{b +\foo{L}_{\dset{z}}(\thetav)}{a})\diff Q(\thetav)
& < & \int \dot{f}^{-1}(- \frac{b + \delta^\star_{Q, \dset{z}}}{a})\diff Q(\thetav)
\label{Eq_ProofKbarNoLambdaIsFinite_pf}\\
& = & \dot{f}^{-1}(- \frac{b + \delta^\star_{Q, \dset{z}}}{a})\\
& < & \infty,\label{Eq_ProofKbarNoLambdaIsFinite_pf_s3}
\end{IEEEeqnarray}
where~\eqref{Eq_ProofKbarNoLambdaIsFinite_pf_s3} follows from $\set{A}_{Q, \dset{z}} \subseteq (0,\infty)$, which implies $a > 0$.
%
For all $(a_1, a_2) \in \set{A}_{Q, \dset{z}}^2$, such that $a_1 < a < a_2$, it holds that for all $\thetav \in \supp Q$,
\begin{IEEEeqnarray}{rCCCl}
	-\frac{1}{a_1}(\foo{L}_{\dset{z}}(\thetav) + b) 
	& < &-\frac{1}{a}(\foo{L}_{\dset{z}}(\thetav) + b) 
	& < & -\frac{1}{a_2}(\foo{L}_{\dset{z}}(\thetav) + b),
\end{IEEEeqnarray}
which from Lemma~\ref{lemm_f_invIsInc} implies that
\begin{IEEEeqnarray}{rCCCl}
\label{EqIneq_cnvxcnj_appc}
	\dot{f}^{-1}(-\frac{1}{a_1}(\foo{L}_{\dset{z}}(\thetav) + b) )  & < & \dot{f}^{-1}(-\frac{1}{a}(\foo{L}_{\dset{z}}(\thetav) + b) )  & < & \dot{f}^{-1}(-\frac{1}{a_2}(\foo{L}_{\dset{z}}(\thetav) + b) ).
\end{IEEEeqnarray}
From~\eqref{EqIneq_cnvxcnj_appc}, it holds that
\begin{IEEEeqnarray}{rCl}
	1 & = & \int  \dot{f}^{-1}(-\frac{1}{a}(\foo{L}_{\dset{z}}(\thetav) + b) ) \diff Q(\thetav)\\
	  & > & \int \dot{f}^{-1}(-\frac{1}{a_1}(\foo{L}_{\dset{z}}(\thetav) + b) ) \diff Q(\thetav).
\end{IEEEeqnarray}
Similarly, from~\eqref{EqIneq_cnvxcnj_appc}, it holds that
\begin{IEEEeqnarray}{rCl}
	1 & = & \int  \dot{f}^{-1}(-\frac{1}{a}(\foo{L}_{\dset{z}}(\thetav) + b) ) \diff Q(\thetav)\\
	  & < & \int \dot{f}^{-1}(-\frac{1}{a_2}(\foo{L}_{\dset{z}}(\thetav) + b) ) \diff Q(\thetav).
\end{IEEEeqnarray}
Then, for the $N_{Q, \dset{z}}(a_1)$ to satisfy,
\begin{IEEEeqnarray}{rCl}
	\int  \dot{f}^{-1}(-\frac{1}{a_1}(\foo{L}_{\dset{z}}(\thetav) + N_{Q, \dset{z}}(a_1))) \diff Q(\thetav) & = & 1,
\end{IEEEeqnarray}
and for the $N_{Q, \dset{z}}(a_2)$ to satisfy,
\begin{IEEEeqnarray}{rCl}
	\int  \dot{f}^{-1}(-\frac{1}{a_2}(\foo{L}_{\dset{z}}(\thetav) + N_{Q, \dset{z}}(a_2))) \diff Q(\thetav) & = & 1,
\end{IEEEeqnarray}
under the assumption that $a_1 < a < a_2$, it holds that $N_{Q, \dset{z}}(a_1)$ and $N_{Q, \dset{z}}(a_2)$ satisfy
\begin{IEEEeqnarray}{rCCCl}
\label{Eq_ProofKbarStricIncreaseGam1vs2}
N_{Q, \dset{z}}(a_1)  & < & b & < & 	N_{Q, \dset{z}}(a_2),
\end{IEEEeqnarray}
which implies that the function $N_{Q, \dset{z}}$ in~\eqref{EqDefNormFunction} is strictly increasing.

For all $(b_1, b_2) \in \set{B}_{Q, \dset{z}}^2$, such that $b_1 < b < b_2$, it holds that for all $\thetav \in \supp Q$,
\begin{IEEEeqnarray}{rCCCl}
	-\frac{1}{a}(\foo{L}_{\dset{z}}(\thetav) + b_1) 
	& > &-\frac{1}{a}(\foo{L}_{\dset{z}}(\thetav) + b) 
	& > & -\frac{1}{a}(\foo{L}_{\dset{z}}(\thetav) + b_2),
\end{IEEEeqnarray}
which from Lemma~\ref{lemm_f_invIsInc} implies that
\begin{IEEEeqnarray}{rCCCl}
\label{EqIneq_cnvxcnj_2_appc}
	\dot{f}^{-1}(-\frac{1}{a}(\foo{L}_{\dset{z}}(\thetav) + b_1) )  & > & \dot{f}^{-1}(-\frac{1}{a}(\foo{L}_{\dset{z}}(\thetav) + b) )  & > & \dot{f}^{-1}(-\frac{1}{a}(\foo{L}_{\dset{z}}(\thetav) + b_2) ).
\end{IEEEeqnarray}
From~\eqref{EqIneq_cnvxcnj_2_appc}, it holds that
\begin{IEEEeqnarray}{rCl}
	1 & = & \int  \dot{f}^{-1}(-\frac{1}{a}(\foo{L}_{\dset{z}}(\thetav) + b) ) \diff Q(\thetav)\\
	  & < & \int \dot{f}^{-1}(-\frac{1}{a}(\foo{L}_{\dset{z}}(\thetav) + b_1) ) \diff Q(\thetav).
\end{IEEEeqnarray}
Similarly, from~\eqref{EqIneq_cnvxcnj_2_appc}, it holds that
\begin{IEEEeqnarray}{rCl}
	1 & = & \int  \dot{f}^{-1}(-\frac{1}{a}(\foo{L}_{\dset{z}}(\thetav) + b) ) \diff Q(\thetav)\\
	  & > & \int \dot{f}^{-1}(-\frac{1}{a}(\foo{L}_{\dset{z}}(\thetav) + b_2) ) \diff Q(\thetav).
\end{IEEEeqnarray}
Then, for the $a_1$ to satisfy,
\begin{IEEEeqnarray}{rCl}
	\int  \dot{f}^{-1}(-\frac{1}{a_1}(\foo{L}_{\dset{z}}(\thetav) + b_1)) \diff Q(\thetav) & = & 1,
\end{IEEEeqnarray}
and for the $a_2$ to satisfy,
\begin{IEEEeqnarray}{rCl}
	\int  \dot{f}^{-1}(-\frac{1}{a_2}(\foo{L}_{\dset{z}}(\thetav) + b_2)) \diff Q(\thetav) & = & 1,
\end{IEEEeqnarray}
under the assumption that $b_1 < b < b_2$, it holds that $a_1$ and $a_2$ satisfy 
\begin{IEEEeqnarray}{rCCCl}
\label{Eq_ProofKbarStricIncreaseGam1vs2_2}
a_1  & < & a & < & 	a_2,
\end{IEEEeqnarray}
which implies that the function $N_{Q, \dset{z}}$ in~\eqref{EqDefNormFunction} is strictly increasing.
Furthermore, from~\eqref{Eq_ProofKbarStricIncreaseGam1vs2} and~\eqref{Eq_ProofKbarStricIncreaseGam1vs2_2} the function $N_{Q, \dset{z}}$ maps one to one for all elements of $\set{A}_{Q, \dset{z}}$ into the $\set{B}_{Q, \dset{z}}$, which implies it is bijective. Thus, the inverse $N^{-1}_{Q, \dset{z}}:\set{B}_{Q, \dset{z}}\to \set{A}_{Q, \dset{z}}$ is well-defined for all $b \in \set{B}_{Q, \dset{z}}$, such that $N^{-1}_{Q, \dset{z}}(b) = a$.
This completes the proof of the first part. 

In the second part, the objective is to prove the continuity of the function $N_{Q, \vect{z}}$. To do so, an auxiliary function is introduced and proven to be continuous.
%
%
Under the assumptions \ref{assum:a}, \ref{assum:b} and \ref{assum:c} from Theorem~\ref{Theo_f_ERMRadNik}, the sets $\set{A}_{Q, \dset{z}}$ and $\set{B}_{Q, \dset{z}}$ in~\eqref{EqDefNormFunction} are non-empty such that 
\begin{IEEEeqnarray}{rCl}
	\bar{a} & = & \sup \set{A}_{Q, \dset{z}},\\
	\underline{a} & = & \inf \set{A}_{Q, \dset{z}},\\
	\bar{b} & = & \sup \set{B}_{Q, \dset{z}}, \text{ and}\\
	\underline{b} & = & \inf \set{B}_{Q, \dset{z}},
\end{IEEEeqnarray}
such that 
\begin{subequations}
\label{EqDefABforF}
\begin{IEEEeqnarray}{rCl}
	\set{A} & = & (\underline{a},\bar{a}) \subseteq (0,\infty), \text{ and} \\
	\set{B} & = & (\underline{b},\bar{b}) \subseteq \reals.
\end{IEEEeqnarray}
\end{subequations}
Let the function $F: \set{A}\times \set{B} \to (0,\infty)$ be
\begin{IEEEeqnarray}{rcl}
\label{EqkWAWALem5}
F(a,b) & = &  \int \dot{f}^{-1}(-\frac{b + \foo{L}_{\dset{z}}(\thetav)}{a}) \diff Q (\vect{\theta})-1.
\end{IEEEeqnarray}
The first step is to prove that the functions $F$ in~\eqref{EqkWAWALem5} is continuous in $\set{A}$ and $\set{B}$ defined in~\eqref{EqDefABforF}, respectively. This is proved by showing that $F$ always exhibits a limit in $\set{A}$ and $\set{B}$.
Then, for all $(a,b) \in \set{A}\times \set{B}$ and for all $\vect{\theta} \in \supp Q$, it holds that 
\begin{IEEEeqnarray}{rcl}
\label{EqThoseEyesISawToday}
\dot{f}^{-1}(\frac{-b - \foo{L}_{\dset{z}}(\thetav)}{a}) & \leq  & \dot{f}^{-1}(-\frac{b + \delta^\star_{Q, \dset{z}}}{a}) < \infty, 
\end{IEEEeqnarray}
where equality holds if and only if $ \foo{L}_{\dset{z}}(\thetav)  = \delta^\star_{Q, \dset{z}}$.
Now, from Corollary~24.5.1 in~\cite{rockafellar1970conjugate} the function $\dot{f}^{-1}$ is continuous, such that for all $b \in \set{B}$, it holds that
\begin{IEEEeqnarray}{rcl}
\label{EqLimRNparB}
\lim_{b \to \beta} \dot{f}^{-1}(\frac{-b - \foo{L}_{\dset{z}}(\thetav)}{a}) & = &  \dot{f}^{-1}(\frac{-\beta - \foo{L}_{\dset{z}}(\thetav)}{a}).
\end{IEEEeqnarray}
Hence, from the dominated convergence theorem by~\cite{ash2000probability} in Theorem~$1.6.9$, the following limit exists and satisfies
\begin{IEEEeqnarray}{rcl}
\label{EqLimContFbeta}
\lim_{b \to \beta} F(a,b) & = & \lim_{b \to \beta}  \int \dot{f}^{-1}(-\frac{b + \foo{L}_{\dset{z}}(\thetav)}{a} ) \diff Q (\vect{\theta})-1\\
& = &  \int  (\lim_{b \to \beta} \dot{f}^{-1}(-\frac{b   + \foo{L}_{\dset{z}}(\thetav)}{a} ) )\diff Q (\vect{\theta})-1\\
& = & \int  \dot{f}^{-1}(-\frac{\beta + \foo{L}_{\dset{z}}(\thetav)}{a}  )\diff Q (\vect{\theta})-1\\
& = & F(\beta,a),
\end{IEEEeqnarray}
which proves that the function $F$ in~\eqref{EqkWAWALem5} is continuous in $\set{B}$.
Similarly, from Corollary~24.5.1 in~\cite{rockafellar1970conjugate} the function $\dot{f}^{-1}$ is continuous, such that for all $a \in \set{A}$, it holds that
\begin{IEEEeqnarray}{rcl}
\label{EqLimRNparA}
\lim_{a \to \lambda} \dot{f}^{-1}(\frac{-b - \foo{L}_{\dset{z}}(\thetav)}{a}) & = &  \dot{f}^{-1}(\frac{-b - \foo{L}_{\dset{z}}(\thetav)}{\lambda}).
\end{IEEEeqnarray}
Hence, from the dominated convergence theorem by~\cite{ash2000probability} in Theorem~$1.6.9$, the following limit exists and satisfies
\begin{IEEEeqnarray}{rcl}
\label{EqLimContFlambda}
\lim_{a \to \lambda} F(a,b) & = & \lim_{a \to \lambda}  \int \dot{f}^{-1}(\frac{-b - \foo{L}_{\dset{z}}(\thetav)}{a} ) \diff Q (\vect{\theta})-1\\
& = &  \int  (\lim_{a \to \lambda} \dot{f}^{-1}(\frac{-b - \foo{L}_{\dset{z}}(\thetav)}{a} ) )\diff Q (\vect{\theta})-1\\
& = & \int  \dot{f}^{-1}(\frac{-b - \foo{L}_{\dset{z}}(\thetav)}{\lambda}  )\diff Q (\vect{\theta})-1 \\
& = & F(\lambda,b),
\end{IEEEeqnarray}
which proves that the function $F$ in~\eqref{EqkWAWALem5} is continuous in $\set{A}$.
The proof continues by noting that from the definition of $\set{A}$ and $\set{B}$ in~\eqref{EqDefABforF} there exists at least one point $(\lambda, \beta) \in \set{A}\times\set{B}$, such that 
\begin{IEEEeqnarray}{rcl}
 (\lambda, \beta) \in \set{A}_{Q,\dset{z}}\times \set{B}_{Q,\dset{z}}, 
\end{IEEEeqnarray}
which implies that
\begin{IEEEeqnarray}{rcl}
 F(\lambda,\beta) 
 	& = &  \int \dot{f}^{-1}(-\frac{\beta + \foo{L}_{\dset{z}}(\thetav)}{\lambda} ) \diff Q (\vect{\theta})-1\\
 	& = & \int \frac{\diff \Pgibbs{P}{Q}}{\diff Q}(\thetav) \diff Q (\vect{\theta})-1\\
 	& = & \int \diff \Pgibbs{P}{Q} (\vect{\theta})-1\\
 	& = & 0.
\end{IEEEeqnarray}
Note that from~\eqref{EqLimContFbeta} and~\eqref{EqLimContFlambda} the function $F$ is continuous and thus the partial derivative of $F$ satisfy
\begin{IEEEeqnarray}{rcl}
 \label{EqDeffaFab}
 \frac{\partial}{\partial a}F(a,b) 
 	& = &  \frac{\partial}{\partial a} (\int \dot{f}^{-1}(-\frac{b + \foo{L}_{\dset{z}}(\thetav)}{a} ) \diff Q (\vect{\theta})-1)\\
 	& = & \int \frac{\partial}{\partial a} \dot{f}^{-1}(-\frac{b + \foo{L}_{\dset{z}}(\thetav)}{a} ) \diff Q (\vect{\theta})\\
 	& = & \int \frac{\diff }{\diff a} \dot{f}^{-1}(-\frac{b + \foo{L}_{\dset{z}}(\thetav)}{a} ) \diff Q (\vect{\theta})\\
 	& = & \int \frac{b + \foo{L}_{\dset{z}}(\thetav)}{a^{2}}\frac{1}{\ddot{f}(\dot{f}^{-1}(-\frac{b + \foo{L}_{\dset{z}}(\thetav)}{a}))} \diff Q (\vect{\theta}), \label{EqDeffaFab_s4}
\end{IEEEeqnarray}
where~\eqref{EqDeffaFab_s4} follows from Lemma~\ref{lemm_f_NormInv}; and
\begin{IEEEeqnarray}{rcl}
 \label{EqDeffbFab}
 \frac{\partial}{\partial b}F(a,b) 
 	& = &  \frac{\partial}{\partial b} (\int \dot{f}^{-1}(-\frac{b + \foo{L}_{\dset{z}}(\thetav)}{a} ) \diff Q (\vect{\theta})-1)\\
 	& = & \int \frac{\partial}{\partial b} \dot{f}^{-1}(-\frac{b + \foo{L}_{\dset{z}}(\thetav)}{a} ) \diff Q (\vect{\theta})\\
 	& = & \int \frac{\diff}{\diff b} \dot{f}^{-1}(-\frac{b + \foo{L}_{\dset{z}}(\thetav)}{a} ) \diff Q (\vect{\theta})\\
 	& = & \int -\frac{1}{a}\frac{1}{\ddot{f}(\dot{f}^{-1}(-\frac{b + \foo{L}_{\dset{z}}(\thetav)}{a}))} \diff Q (\vect{\theta}),\label{EqDeffbFab_s4}
\end{IEEEeqnarray}
where~\eqref{EqDeffbFab_s4} follows from Lemma~\ref{lemm_f_NormInv}.
Then, from \emph{The Implicit Function Theorem} presented in~\cite[Theorem 4]{oswaldo2013TIFT}, the function~$N_{Q, \dset{z}}$ exists and is unique in the open interval $\set{A}$ with $\set{A}$ in~\eqref{EqDefABforF} and for all $a \in \set{A}$ satisfies that
\begin{IEEEeqnarray}{rcl}
N_{Q, \dset{z}}(a) & = &  b, 
\end{IEEEeqnarray}
such that 
\begin{IEEEeqnarray}{rcl}
F(a,N_{Q, \dset{z}}(a)) & = &  0, 
\end{IEEEeqnarray}
which completes the proof of continuity for the normalization function $N_{Q, \dset{z}}$.
Additionally, from Theorem 4 in~\cite{oswaldo2013TIFT} it follows that 
\begin{IEEEeqnarray}{rcl}
\frac{\diff }{\diff a}N_{Q, \dset{z}}(a) 
& = & -  (\frac{\partial}{\partial b}F(a,N_{Q, \dset{z}}(a)))^{-1}  \frac{\partial}{\partial a}F(a,N_{Q, \dset{z}}(a)),\\
& = &-\frac{\displaystyle\int \frac{N_{Q, \dset{z}}(a) + \foo{L}_{\dset{z}}(\thetav)}{a^{2}}\frac{1}{\ddot{f}(\dot{f}^{-1}(-\frac{N_{Q, \dset{z}}(a) + \foo{L}_{\dset{z}}(\thetav)}{a}))} \diff Q (\thetav)}{\displaystyle\int -\frac{1}{a}\frac{1}{\ddot{f}(\dot{f}^{-1}(-\frac{N_{Q, \dset{z}}(a) + \foo{L}_{\dset{z}}(\nuv)}{a}))} \diff Q (\nuv)} \\
& = &\frac{\displaystyle\int \frac{N_{Q, \dset{z}}(a) + \foo{L}_{\dset{z}}(\thetav)}{a}\frac{1}{\ddot{f}(\dot{f}^{-1}(-\frac{N_{Q, \dset{z}}(a) + \foo{L}_{\dset{z}}(\thetav)}{a}))} \diff Q (\thetav)}{\displaystyle\int \frac{1}{\ddot{f}(\dot{f}^{-1}(-\frac{N_{Q, \dset{z}}(a) + \foo{L}_{\dset{z}}(\nuv)}{a}))} \diff Q (\nuv)} \\
& = &\frac{\displaystyle\int \frac{N_{Q, \dset{z}}(a) + \foo{L}_{\dset{z}}(\thetav)}{a}(\ddot{f}(\frac{\diff \Pgibbs[\dset{z}][a]{P}{Q}}{\diff Q}(\thetav)))^{-1} \diff Q (\thetav)}{\displaystyle\int (\ddot{f}(\frac{\diff \Pgibbs[\dset{z}][a]{P}{Q}}{\diff Q}(\nuv)))^{-1} \diff Q (\nuv)} \\
& = &\frac{N_{Q, \dset{z}}(a)}{a}+ \frac{1}{a}\frac{\displaystyle\int\foo{L}_{\dset{z}}(\thetav)(\ddot{f}(\frac{\diff \Pgibbs[\dset{z}][a]{P}{Q}}{\diff Q}(\thetav)))^{-1} \diff Q (\thetav)}{\displaystyle\int (\ddot{f}(\frac{\diff \Pgibbs[\dset{z}][a]{P}{Q}}{\diff Q}(\nuv)))^{-1} \diff Q (\nuv)}\\
& = &\frac{N_{Q, \dset{z}}(a)}{a}+ \frac{1}{a}\frac{\displaystyle\int\foo{L}_{\dset{z}}(\thetav)(\ddot{f}(\frac{\diff \Pgibbs[\dset{z}][a]{P}{Q}}{\diff Q}(\thetav)))^{-1} \diff Q (\thetav)}{\displaystyle\int (\ddot{f}(\frac{\diff \Pgibbs[\dset{z}][a]{P}{Q}}{\diff Q}(\nuv)))^{-1} \diff Q (\nuv)}\\
& = & \frac{N_{Q, \dset{z}}(a)}{a}+ \frac{1}{a}\displaystyle\int\foo{L}_{\dset{z}}(\thetav)\frac{\frac{1}{\displaystyle\ddot{f}(\frac{\diff \Pgibbs[\dset{z}][a]{P}{Q}}{\diff Q}(\thetav))}}{\displaystyle\int \frac{1}{\ddot{f}(\frac{\diff \Pgibbs[\dset{z}][a]{P}{Q}}{\diff Q}(\nuv))} \diff Q (\nuv)} \diff Q (\thetav).\label{EqProofNzFuncDef_s7}
\end{IEEEeqnarray}

The proof continues by considering a function $g_a:\set{M} \to \reals$, such that for all $\thetav \in \supp Q$
\begin{IEEEeqnarray}{rcl}
\label{EqRNnewRm}
	g_a(\thetav) & = & \frac{\frac{1}{\displaystyle\ddot{f}(\frac{\diff \Pgibbs[\dset{z}][a]{P}{Q}}{\diff Q}(\thetav))}}{\displaystyle\int \frac{1}{\ddot{f}(\frac{\diff \Pgibbs[\dset{z}][a]{P}{Q}}{\diff Q}(\nuv))} \diff Q (\nuv)}.
\end{IEEEeqnarray}
Note that from the assumption that $f$ is strictly convex and twice differentiable, the derivative $\dot{f}$ is increasing, and the second derivative $\ddot{f}$ is positive for all $\thetav \in \supp Q$. Also, the denominator of the fraction is the integral of the reciprocal of $\ddot{f}\left( \frac{\diff \Pgibbs[\dset{z}][a]{P}{Q}}{\diff Q}(\nuv) \right)$ with respect to the measure $Q$. This term serves as a normalization constant ensuring that the resulting function is a proper probability density such that
\begin{IEEEeqnarray}{rcl}
	\int g_a(\thetav) \diff Q (\thetav)& = & 1.
\end{IEEEeqnarray}
Therefore, the function $g_a$ in~\eqref{EqRNnewRm} can be interpreted as the Radon-Nikodym derivative of a new probability measure $P^{(a)}$, parametrizes by the regularization factor $a$ with respect to $Q$. Specifically, if we define a measure $P^{(a)}$ such that for any set $\set{A} \in \field{F}_{\set{M}}$,
\begin{IEEEeqnarray}{rcl}
	P^{(a)}(\set{A}) = \int_{\set{A}} \frac{\frac{1}{\ddot{f}\left( \frac{\diff \Pgibbs[\dset{z}][a]{P}{Q}}{\diff Q}(\thetav) \right)}}{\displaystyle\int \frac{1}{\ddot{f}\left( \frac{\diff \Pgibbs[\dset{z}][a]{P}{Q}}{\diff Q}(\nuv) \right)} \, \diff Q (\nuv)} \, dQ(\thetav).
\end{IEEEeqnarray}
Therefore, for all $\thetav \in \supp Q$ it follows that 
\begin{IEEEeqnarray}{rcl}
\label{EqGaisRNnewRm}
	g_a(\thetav) & = & \frac{\diff P^{(a)}}{\diff Q}(\thetav).
\end{IEEEeqnarray}
From~\eqref{EqProofNzFuncDef_s7} and~\eqref{EqGaisRNnewRm}
\begin{IEEEeqnarray}{rcl}
	N_{Q, \dset{z}}(a) 
	& = & a \frac{\diff }{\diff a}N_{Q, \dset{z}}(a) -\displaystyle\int\foo{L}_{\dset{z}}(\thetav)\frac{\diff P^{(a)}}{\diff Q}(\thetav) \diff Q (\thetav)\\
	& = & a \frac{\diff }{\diff a}N_{Q, \dset{z}}(a) - \foo{R}_{\dset{z}}(P^{(a)}),
\end{IEEEeqnarray}
with $\foo{R}_{\dset{z}}$ defined in~\eqref{EqRxy}.
This completes the proof of the derivative of the Normalization function.
\end{IEEEproof}
\subsection{Proof of Lemma~\ref{lemm_fDR_kset}}
\label{app_proof_lemm_fDR_kset}
\begin{IEEEproof}
\label{proof_lemm_fDR_kset}
Given a reference measure $Q$, a dataset $\dset{z}$, a strictly convex and differentiable function $f$ that induces an $f$-divergence and the empirical risk function $\foo{L}_{\dset{z}}$ in~\eqref{EqLxy}, under the assumption that there exists a $\lambda \in (0,\infty)$, such that the optimization problem~\eqref{EqOp_f_ERMRERNormal} has a solution, the proof is concerned with characterizing the set of all regularization factors $\lambda$ for which a solution exists. This set of regularization factors is denoted by $\set{A}_{Q, \dset{z}}$, where $\set{A}_{Q, \dset{z}} \subseteq (0,\infty)$.
The proof is divided into three parts. 
In the first part, the Legendre-Fenchel transform of $f$ is connected to the Radon-Nikodym derivative of the solution to the optimization problem in~\eqref{EqOp_f_ERMRERNormal} presented in Theorem~\ref{Theo_f_ERMRadNik}.
In the second part, the strictly increasing property of the Radon-Nikodym derivative, obtained from the connection established with the Legendre-Fenchel transform of $f$, is used to evaluate the real values of $\lambda$ under which assumption~\eqref{EqDefSetB} holds.
In the third part, the strictly increasing property is used to evaluate the real values of $\lambda$ under which assumption~\eqref{EqEqualToABigOne} holds.


The first part is as follows.
The Legendre-Fenchel transform of $f$ is defined as
\begin{IEEEeqnarray}{rCl}
\label{Eq_LFT1_pf}
	f^{*}(t) & \triangleq & \sup_{s\in \set{I}}( ts- f(s)),
\end{IEEEeqnarray} 
where $f^{*}:\set{J}\rightarrow\reals$.
From Assumption~\ref{assum:a} and \cite[Theorem 23.5]{rockafellar1970conjugate} the Legendre-Fenchel transform $f^{*}$ in~\eqref{Eq_LFT1_pf} satisfies
\begin{IEEEeqnarray}{rCl}
\label{Eq_convcnj2_pf}
	f^{*}(t) & = &  t\frac{\diff }{\diff t}f^{*}(t)- f(\frac{\diff }{\diff t}f^{*}(t)).
\end{IEEEeqnarray}
Furthermore, from Corollary 23.5.1 in~\cite{rockafellar1970conjugate}, the function $\frac{\diff }{\diff t}f^{*}:\set{J}\rightarrow \set{I}$ satisfies 
\begin{IEEEeqnarray}{rCl}
\label{Eq_convcnj4_pf}
	\frac{\diff }{\diff t}f^{*}(t) & = & ({\frac{\diff f}{\diff t}})^{-1}(t),
\end{IEEEeqnarray}
which is the functional inverse of the derivative of $f$, denoted by $\dot{f}^{-1}$ for simplicity.
Note that, given the assumption that $f$ is strictly convex and induces an $f$-divergence, it follows from Theorem 12.2 in~\cite{rockafellar1970conjugate} that the function $f^{*}$ in~\eqref{Eq_LFT1_pf} is also strictly convex.
From the strict convexity of $f^{*}$, it follows from Lemma~\ref{lemm_f_invIsInc} that $\dot{f}^{-1}$ in~\eqref{Eq_convcnj4_pf} is strictly increasing.
Furthermore, from Corollary 26.3.1 in~\cite{rockafellar1970conjugate} $f^{*}$ in~\eqref{Eq_LFT1_pf} is bijective with $\frac{\diff f}{\diff s}:\set{I}\rightarrow \set{J}$, which completes the first part of the proof.


The second part is as follows.
Evaluating the real values of $\lambda$ under which assumption in~\eqref{EqDefSetB} holds requires to show that the function $\frac{\diff \Pgibbs{P}{Q}}{\diff Q}$ belongs to the set of nonnegative measurable functions.
From the $f$-divergence in Definition~\ref{Def_fDivergence} and the fact that $\dot{f}^{-1}$ strictly increasing and bijective, the proof follows by showing that the limit 
\begin{IEEEeqnarray}{rCl}
\label{Eq_limdf_at_0}
	\lim_{x\rightarrow 0^{+}}\dot{f}(x) & = & t_0,
\end{IEEEeqnarray}
satisfies for all $\thetav \in \supp Q$,
\begin{IEEEeqnarray}{rCl}
\label{Eq_limdf_argument}
	-\frac{\foo{L}_{\dset{z}}(\thetav)+\beta}{\lambda} & > & t_0.
\end{IEEEeqnarray}
Note that~\eqref{Eq_limdf_argument} is sufficient from the fact that the monotonicity of $\dot{f}^{-1}$ implies that for all $t > t_0$, 
\begin{IEEEeqnarray}{rCl}
	\dot{f}^{-1}(t) & > & 0.
\end{IEEEeqnarray}
To evaluate the real values of $\lambda$ under which assumption in~\eqref{EqDefSetB} holds, three cases must be considered for the limit in~\eqref{Eq_limdf_at_0}.

{\bf Case 1:} Assume that
\begin{IEEEeqnarray}{rCl}
\label{Eq_limdf_at_0_inf}
	\lim_{x\rightarrow 0^{+}}\dot{f}(x) & = & \infty.
\end{IEEEeqnarray}
Under the above assumption, for all $\thetav \in \supp Q$,
\begin{IEEEeqnarray}{rCl}
\label{Eq_limdf_argumentNot}
	-\frac{\foo{L}_{\dset{z}}(\thetav)+\beta}{\lambda} & < & \infty,
\end{IEEEeqnarray}
which implies that
\begin{IEEEeqnarray}{rCl}
	\dot{f}^{-1}(-\frac{\foo{L}_{\dset{z}}(\thetav)+\beta}{\lambda}) & < & 0.
\end{IEEEeqnarray}
Hence, Assumption~\ref{assum:b} in Theorem~\ref{Theo_f_ERMRadNik} is not satisfied and nothing can be stated about the solution.

{\bf Case 2:} Assume that
\begin{IEEEeqnarray}{rCl}
\label{Eq_limdf_at_0_a}
	\lim_{x\rightarrow 0^{+}}\dot{f}(x) & = & a,
\end{IEEEeqnarray}
where $a \in \reals$. Under the above assumption, consider the set
\begin{IEEEeqnarray}{rCl}
\label{Eq_limdf_at_0_a_set}
	\set{D} & = & \{\thetav \in \supp Q: -\foo{L}_{\dset{z}}(\thetav) < a\lambda + \beta\}.
\end{IEEEeqnarray}
On one hand, note that if the function $\foo{L}_{\dset{z}}$ in~\eqref{EqLxy} is unbounded in $\supp Q$, from~\eqref{Eq_limdf_at_0_a} the set $\set{D}$ in~\eqref{Eq_limdf_at_0_a_set} is nonegligble and measurable, such that for all $\thetav \in \set{D}$,
\begin{IEEEeqnarray}{rCl}
	-\frac{\foo{L}_{\dset{z}}(\thetav)+\beta}{\lambda} & < & a,
\end{IEEEeqnarray}
which implies that 
\begin{IEEEeqnarray}{rCl}
	\dot{f}^{-1}(-\frac{\foo{L}_{\dset{z}}(\thetav)+\beta}{\lambda}) & < & 0.
\end{IEEEeqnarray}
Hence, Assumption~\ref{assum:b} in Theorem~\ref{Theo_f_ERMRadNik} is not satisfied and nothing can be stated about the solution.
On the other hand, if the function $\foo{L}_{\dset{z}}$ in~\eqref{EqLxy} is bounded in $\supp Q$, such that
\begin{IEEEeqnarray}{rCl}
\label{Eq_supMcase2}
	M & = & \sup_{\thetav \in \supp Q} \foo{L}_{\dset{z}}(\thetav).
\end{IEEEeqnarray}
Then, there exists a $\lambda_{Q, \dset{z}} \in (0,\infty)$ such that
\begin{IEEEeqnarray}{rCl}
\label{EqlambdaStarp1}
	-M & = & a\lambda_{Q, \dset{z}}+\beta.
\end{IEEEeqnarray}
From~\eqref{EqlambdaStarp1} for all $\lambda > \lambda_{Q, \dset{z}}$, it holds that for all $\thetav \in \supp Q$,
\begin{IEEEeqnarray}{rCl}
\label{EqconstNonegCase2}
	\dot{f}^{-1}(-\frac{\foo{L}_{\dset{z}}(\thetav)+\beta}{\lambda}) & > & 0.
\end{IEEEeqnarray}
From~\eqref{EqconstNonegCase2}, consider the following conditions:
If there exists a model $\bar{\thetav} \in \supp Q$ such that $\foo{L}_{\dset{z}}(\bar{\thetav}) = M$, where $M$ is defined in~\eqref{Eq_supMcase2}, then the set of regularization factors $\lambda$ for which the function $\frac{\diff \Pgibbs{P}{Q}}{\diff Q}$ is nonnegative is $[\lambda_{Q, \dset{z}}, \infty)$.
Alternatively, if for all models $\thetav \in \supp Q$, it holds that $\foo{L}_{\dset{z}}(\thetav) < M$, where $M$ is defined in~\eqref{Eq_supMcase2}, then the set of regularization factors $\lambda$ for which the function $\frac{\diff \Pgibbs{P}{Q}}{\diff Q}$ is nonnegative is $(\lambda_{Q, \dset{z}}, \infty)$.

{\bf Case 3:} Assume that
\begin{IEEEeqnarray}{rCl}
\label{Eq_limdf_at_0_ninfty}
	\lim_{x\rightarrow 0^{+}}\dot{f}(x) & = & -\infty.
\end{IEEEeqnarray}
Under the above assumption, for all $\thetav \in \supp Q$,
\begin{IEEEeqnarray}{rCl}
\label{Eq_limdf_argumentTrue}
	-\frac{\foo{L}_{\dset{z}}(\thetav)+\beta}{\lambda} & > & -\infty,
\end{IEEEeqnarray}
which implies that
\begin{IEEEeqnarray}{rCl}
	\dot{f}^{-1}(-\frac{\foo{L}_{\dset{z}}(\thetav)+\beta}{\lambda}) & > & 0.
\end{IEEEeqnarray}
Hence, for all $\lambda \in (0,\infty)$ Assumption~\ref{assum:b} in Theorem~\ref{Theo_f_ERMRadNik} is satisfied such that the nonnegativity of the function $\frac{\diff \Pgibbs{P}{Q}}{\diff Q}$ is guaranteed.
This completes the second part of the proof.


The third part is as follows.
Evaluating the values $\lambda$ under which assumption~\eqref{EqEqualToABigOne} holds requires showing that there exists a real value $\beta \in \reals$ such that the integral of $\frac{\diff \Pgibbs{P}{Q}}{\diff Q}$ with respect to $Q$ is one. 
From Theorem~\ref{theo_InfDevKfDR} the monotonicity of the normalization function $N_{Q, \dset{z}}$ in~\eqref{EqDefNormFunction}, there is a minimum regularization factor $\lambda^{\star}_{Q, \dset{z}}$ defined in~\eqref{EqDefLambdaStar}. Furthermore, from Theorem~\ref{theo_InfDevKfDR} the continuity of the function $N_{Q, \dset{z}}$ implies that for all $\lambda \in (\lambda^{\star}_{Q, \dset{z}}, \infty)$, there exists a unique $\beta \in \set{B}_{Q, \dset{z}}$ such that Assumption~\ref{assum:b} of Theorem~\ref{Theo_f_ERMRadNik} is satisfied.
From Theorem~\ref{theo_InfDevKfDR}, it holds that
\begin{IEEEeqnarray}{rCl}
\lim_{\lambda \rightarrow {\lambda^{\star}_{Q, \dset{z}}}^{+}} N_{Q, \dset{z}}(\lambda) & = & N_{Q, \dset{z}}(\lambda^{\star}_{Q, \dset{z}}),
\end{IEEEeqnarray}
with the function $N_{Q, \dset{z}}$ defined in~\eqref{EqDefNormFunction} and the limit from the right is well-defined from the fact that the set $\set{A}_{Q, \dset{z}}$ is convex. 
To determine whether the infimum in~\eqref{EqDefLambdaStar} belongs to the set $\set{A}_{Q, \dset{z}}$ two cases are considered.

{\bf Case 1:} Assume that $\beta > N_{Q, \dset{z}}(\lambda^{\star}_{Q, \dset{z}})$, such that
\begin{IEEEeqnarray}{rCl}
\label{EqIntAssumption1}
	\int \dot{f}^{-1}(-\frac{\beta + \foo{L}_{\dset{z}}(\thetav)}{\lambda^{\star}_{Q, \dset{z}}})\diff Q(\thetav)
	& = & \infty.
\end{IEEEeqnarray}
Notice that from~\eqref{EqIntAssumption1} for all $\beta_1 \in [ N_{Q, \dset{z}}(\lambda^{\star}_{Q, \dset{z}}), \beta)$ and for all $\thetav \in \supp Q$, it holds that
\begin{IEEEeqnarray}{rCl}
\label{EqIntAssumption1_prt2}
	\dot{f}^{-1}(-\frac{\beta_1 + \foo{L}_{\dset{z}}(\thetav)}{\lambda^{\star}_{Q, \dset{z}}})
	& > & \dot{f}^{-1}(-\frac{\beta + \foo{L}_{\dset{z}}(\thetav)}{\lambda^{\star}_{Q, \dset{z}}}).
\end{IEEEeqnarray}
Hence, under the above assumption, $N_{Q, \dset{z}}(\lambda^{\star}_{Q, \dset{z}}) \notin \set{B}_{Q,\dset{z}}$ which implies that the set of all regularization $\set{A}_{Q, \dset{z}}$ in~\eqref{EqDefNormFunction} that satisfy assumption~\eqref{EqEqualToABigOne} is $\set{A}_{Q, \dset{z}} = (\lambda^{\star}_{Q, \dset{z}},\infty)$.

{\bf Case 2:} Assume that $\beta > N_{Q, \dset{z}}(\lambda^{\star}_{Q, \dset{z}})$, such that
\begin{IEEEeqnarray}{rCl}
\label{EqIntAssumption2}
	\int \dot{f}^{-1}(-\frac{\beta + \foo{L}_{\dset{z}}(\thetav)}{\lambda^{\star}_{Q, \dset{z}}})\diff Q(\thetav)
	& < & \infty.
\end{IEEEeqnarray}
From the monotonicity of the solution in part one and continuity of the function $N_{Q, \dset{z}}$ in~\eqref{EqDefNormFunction} from Theorem~\ref{theo_InfDevKfDR}, there exists a $\beta^{\star}_{Q, \dset{z}} \in \set{B}_{Q,\dset{z}}$  such that $N_{Q, \dset{z}}(\lambda^{\star}_{Q, \dset{z}})=\beta^{\star}_{Q, \dset{z}}$, which implies that the set of all regularization factors $\set{A}_{Q, \dset{z}} = [\lambda^{\star}_{Q, \dset{z}},\infty)$.

Finally, from parts two and three of the proof the set $\set{A}_{Q, \dset{z}}$ is a convex set such that the regularization factors for which the assumptions of Theorem~\ref{Theo_f_ERMRadNik} hold and are given by 
\begin{IEEEeqnarray}{rCl}
	\label{Eq_pfDivfConstrainOpen}
		\set{A}_{Q, \dset{z}} & = & 
		\begin{cases}
 			[\lambda^{\star}_{Q,\dset{z}}, \infty) & \text{if } \displaystyle \int \dot{f}^{-1}(- \frac{\beta+\foo{L}_{\dset{z}}(\thetav)}{\lambda^{\star}_{Q,\dset{z}}})\diff Q(\thetav)< \infty,\vspace{2mm} \\
 			(\lambda^{\star}_{Q,\dset{z}}, \infty)& \text{otherwise},
 		\end{cases}
	\end{IEEEeqnarray}
where $\beta > \lim_{\lambda \rightarrow {\lambda^{\star}_{Q, \dset{z}}}^{+}} N_{Q, \dset{z}}(\lambda)$. This completes the proof.

\end{IEEEproof}
\subsection{Proof of Lemma~\ref{lemm_fDR_No_minRegF_nneg}}
\label{app_proof_lemm_fDR_No_minRegF_nneg}
\begin{IEEEproof}
\label{proof_lemm_fDR_No_minRegF_nneg}
The following proof is divided into two parts. In the first part, an auxiliary function is introduced and proven to be continuous. In the second part, a contradiction is shown under the assumption that $\dot{f}^{-1}$ is nonnegative and the continuity of the auxiliary function. Finally, it is shown that for nonnegative $\dot{f}^{-1}$, the set of admissible regularization factors is the positive reals.

The first part is as follows.
Let the function $k: \reals \to (0, +\infty)$, be such that
\begin{IEEEeqnarray}{rcl}
\label{EqkWA}
k(b) & = &  \int \dot{f}^{-1}(\frac{-b - \foo{L}_{\dset{z}}(\thetav)}{\lambda} ) \diff Q (\vect{\theta}).
\end{IEEEeqnarray}
The first step is to prove that the function $k$ in~\eqref{EqkWA} is continuous in $\reals$. This is proved by showing that $k$ always exhibits a limit. 
Note that from Lemma~\ref{lemm_f_invIsInc} the function $\dot{f}^{-1}$ is strictly increasing, it holds that for all $b \in \set{B}_{Q, \dset{z}}$ with $\set{B}_{Q, \dset{z}}$ defined in~\eqref{EqDefNormFunction} and for all $\vect{\theta} \in \supp Q$, it holds that 
\begin{IEEEeqnarray}{rcl}
\label{EqIDidntSee}
\dot{f}^{-1}(\frac{-b - \foo{L}_{\dset{z}}(\thetav)}{\lambda}) & \leq  & \dot{f}^{-1}(-\frac{b}{\lambda}), \end{IEEEeqnarray}
where equality holds if and only if $ \foo{L}_{\dset{z}}(\thetav)  = 0$.
Now, from the Corollary~24.5.1~\cite{rockafellar1970conjugate} the function $\dot{f}^{-1}$ is continuous, such that for all $a \in \set{B}$, it holds that
\begin{IEEEeqnarray}{rcl}
\label{EqSuchABeautiflFace2}
\lim_{b \to \beta} \dot{f}^{-1}(\frac{-b - \foo{L}_{\dset{z}}(\thetav)}{\lambda}) & = &  \dot{f}^{-1}(\frac{-\beta - \foo{L}_{\dset{z}}(\thetav)}{\lambda}).
\end{IEEEeqnarray}
Hence, from the dominated convergence theorem by~\cite{ash2000probability} in Theorem~1.6.9, the following limit exists and satisfies
\begin{IEEEeqnarray}{rcl}
\label{EqSuchACrazyDay}
\lim_{b \to \beta} k(b) & = & \lim_{b \to \beta}  \int \dot{f}^{-1}(\frac{-b - \foo{L}_{\dset{z}}(\thetav)}{\lambda} ) \diff Q (\vect{\theta})\\
& = &  \int  (\lim_{b \to \beta} \dot{f}^{-1}(\frac{-b   - \foo{L}_{\dset{z}}(\thetav)}{\lambda} ) )\diff Q (\vect{\theta})\\
& = & \int  \dot{f}^{-1}(\frac{-\beta - \foo{L}_{\dset{z}}(\thetav)}{\lambda}  )\diff Q (\vect{\theta})\\
& = & k(\beta),
\end{IEEEeqnarray}
which proves that the function $k$ in~\eqref{EqkWA} is continuous.

The second part is as follows.
From the assumption that $\set{B}_{Q, \dset{z}}$ is nonempty, there is a $b \in \set{B}_{Q, \dset{z}}$ and a $\lambda \in (0,\infty)$ such that,
\begin{IEEEeqnarray}{rCl}
1 = \int \dot{f}^{-1}(-\frac{b+\foo{L}_{\dset{z}}}{\lambda}) \diff Q(\thetav).
\end{IEEEeqnarray}
From Corollary~24.5.1 in~\cite{rockafellar1970conjugate} and Lemma~\ref{lemm_f_invIsInc} the function $\dot{f}^{-1}$ is continuous and strictly increasing, for all $b_1 \in (b^{\star}_{Q,\dset{z}},b)$ and for all $b_2 \in (b,\infty)$, it holds that
\begin{equation}
\label{EqfDRforTs}
\int \dot{f}^{-1}(-\frac{b_1+\foo{L}_{\dset{z}}}{\lambda}) \diff Q(\thetav)  >  1 
>  \int \dot{f}^{-1}(-\frac{b_2+\foo{L}_{\dset{z}}}{\lambda}) \diff Q(\thetav).
\end{equation}
Under the same argument, for all $\lambda_1 \in (0,\lambda)$ and for all $\lambda_2 \in (\lambda,\infty)$, it holds that
\begin{equation}
\label{EqfDRforLambdas}
	\int \dot{f}^{-1}(-\frac{b + \foo{L}_{\dset{z}}}{\lambda_1}) \diff Q(\thetav)  <  1 
	<  \int \dot{f}^{-1}(-\frac{b + \foo{L}_{\dset{z}}}{\lambda_2}) \diff Q(\thetav).
\end{equation}

Hence, given that the function $k$ in~\eqref{EqkWA} is continuous, strictly decreasing, from~\eqref{EqfDRforTs} then, there always exists two reals $b_1$ and $b_2$ in $\set{B}_{Q, \dset{z}}$ such that $k(b_1) < 1 < k(b_2)$, it follows from the intermediate-value in~\cite[Theorem~$4.23$]{rudin1953bookPrinciples} that there always exists a unique real $b \in \set{B}_{Q, \dset{z}}$ such that $k(b) = 1$. Furthermore, for all $b \in \set{B}_{Q, \dset{z}}$ there always exists two reals $\lambda_1$ and $\lambda_2$ in $(0,\infty)$ such that inequality~\eqref{EqfDRforLambdas} holds, it follows from the \emph{intermediate-value theorem}~in~\cite[Theorem~$4.23$]{rudin1953bookPrinciples} that there always exists a unique real $b \in \set{B}_{Q, \dset{z}}$ for all $\lambda \in (0,\infty)$ such that $k(b) = 1$. 
Finally, from the fact that $N_{Q,\dset{z}}$ in~\eqref{EqDefNormFunction} is continuous and strictly increasing, if $\set{B}_{Q, \dset{z}} = (t^{\star}_{Q,\dset{z}},\infty)$ then the set of admissible regularization factors $\set{A}_{Q, \dset{z}}$ in~\eqref{EqDefMapNormFunction} is identical to $(0,\infty)$, which completes the proof.
\end{IEEEproof}

\section{Proofs of Section~\ref{sec:GenErr}}
\subsection{Proof of Theorem \ref{theo_ERM_fDR_LT_AnyP}}
\begin{IEEEproof}
\label{app_theo_ERM_fDR_LT_AnyP}
From Theorem~\ref{Theo_f_ERMRadNik}, Theorem~\ref{Theo_dual_is_N} and Lemma~\ref{lemm_f_LFTequality}, the Legendre-Fenchel transform in Definition~\ref{DefLT_cnvxcnj} satisfies for all $\thetav \in \supp Q$,
\begin{IEEEeqnarray}{rCl}
\label{EqLF_transf}
    f(\frac{\diff \Pgibbs{P}{Q}}{\diff Q}(\thetav))& = & -\frac{\foo{L}_{\dset{z}}(\thetav)+ N_{Q, \dset{z}}(\lambda)}{\lambda}\frac{\diff \Pgibbs{P}{Q}}{\diff Q}(\thetav)-f^{\star}(\frac{\foo{L}_{\dset{z}}(\thetav)+N_{Q, \dset{z}}(\lambda)}{\lambda}),
\end{IEEEeqnarray}
where~\eqref{EqLF_transf} can be rearranged into
\begin{IEEEeqnarray}{rCl}
\label{EqLF_transf2}
     -\frac{\foo{L}_{\dset{z}}(\thetav)+N_{Q, \dset{z}}(\lambda)}{\lambda}\frac{\diff \Pgibbs{P}{Q}}{\diff Q}(\thetav)& = &f(\frac{\diff \Pgibbs{P}{Q}}{\diff Q}(\thetav))+f^{\star}(\frac{\foo{L}_{\dset{z}}(\thetav)+N_{Q, \dset{z}}(\lambda)}{\lambda}).
\end{IEEEeqnarray}

Using~\eqref{EqLF_transf2}, it can be shown that
\begin{IEEEeqnarray}{rCl}
\IEEEeqnarraymulticol{3}{l}{
    \lambda\int ( f(\frac{\diff \Pgibbs{P}{Q}}{\diff Q}(\thetav))+ f^{*}(-\frac{\foo{L}_{\dset{z}}(\thetav)+N_{Q, \dset{z}}(\lambda)}{\lambda}))(1 - \frac{\diff P}{\diff \Pgibbs{P}{Q}}(\thetav)) \diff Q(\thetav)
     } \nonumber \\ 
& = &  \lambda\int  -\frac{\foo{L}_{\dset{z}}(\thetav)+N_{Q, \dset{z}}(\lambda)}{\lambda}\frac{\diff \Pgibbs{P}{Q}}{\diff Q}(\thetav)(1 - \frac{\diff P}{\diff \Pgibbs{P}{Q}}(\thetav)) \diff Q(\thetav)\label{EqGap_Gibb2P_s1}\\
& = &  \lambda\int  \frac{\foo{L}_{\dset{z}}(\thetav)+N_{Q, \dset{z}}(\lambda)}{\lambda}(\frac{\diff P}{\diff Q}(\thetav) - \frac{\diff \Pgibbs{P}{Q}}{\diff Q}(\thetav) ) \diff Q(\thetav)\label{EqGap_Gibb2P_s2}\\
& = &  \lambda\int  \frac{\foo{L}_{\dset{z}}(\thetav)+N_{Q, \dset{z}}(\lambda)}{\lambda}\diff P(\thetav) -\lambda\int \frac{\foo{L}_{\dset{z}}(\thetav)+N_{Q, \dset{z}}(\lambda)}{\lambda} \diff \Pgibbs{P}{Q}(\thetav)\label{EqGap_Gibb2P_s3}\\
& = &  \int \foo{L}_{\dset{z}}(\thetav)\diff P(\thetav) - \int \foo{L}_{\dset{z}}(\thetav)\diff \Pgibbs{P}{Q}(\thetav) +N_{Q, \dset{z}}(\lambda)-N_{Q, \dset{z}}(\lambda)\label{EqGap_Gibb2P_s4}\\
& = & \foo{R}_{\dset{z}}(P) - \foo{R}_{\dset{z}}(\Pgibbs{P}{Q}),\label{EqGap_Gibb2P_s5}\\
& = & \foo{G}(\dset{z},P,\Pgibbs{P}{Q}),\label{EqGap_Gibb2P_s6}
\end{IEEEeqnarray}
where~\eqref{EqGap_Gibb2P_s1} follows from~\eqref{EqLF_transf2},~\eqref{EqGap_Gibb2P_s2} follows from the fact that $P$ is  absolutely continuous with respect to $Q$; and~\eqref{EqGap_Gibb2P_s2} follows from~\eqref{EqGap}.
This completes the proof.
\end{IEEEproof}

\subsection{Proof Theorem~\ref{Theo_ERM_fDR_LT_GenERr}}
\begin{IEEEproof}
\label{app_theo_ERM_fDR_LT_GenERr}
From~\eqref{EqGap} the gap for an arbitrary dataset $\dset{z}$ and two arbitrary probability measures $P_1$ and $P_2$ satisfies
\begin{IEEEeqnarray}{rCl}
	\foo{G}(\dset{z},P_1,P_2) & = &  \foo{R}_{\dset{z}}(P_1) - \foo{R}_{\dset{z}}(P_2)
\end{IEEEeqnarray}
From Theorem~\ref{theo_ERM_fDR_LT_AnyP} the gap for an arbitrary dataset $\dset{z}$ and two arbitrary probability measures $P_1$ and $P_2$ in $\bigtriangleup_Q(\set{M})$, is given satisfies
\begin{IEEEeqnarray}{rCl}
\foo{G}(\dset{z},P_1,P_2) & = &  \foo{G}(\dset{z},P_1,\Pgibbs{P}{Q})-\foo{G}(\dset{z},P_2,\Pgibbs{P}{Q})\\
& = &  \lambda\int ( f(\frac{\diff \Pgibbs{P}{Q}}{\diff Q}(\thetav))+ f^{*}(-\frac{\foo{L}_{\dset{z}}(\thetav)+\beta}{\lambda}))(1 - \frac{\diff P_1}{\diff \Pgibbs{P}{Q}}(\thetav)) \diff Q(\thetav) 
\nonumber\\ &  &
- \lambda\int ( f(\frac{\diff \Pgibbs{P}{Q}}{\diff Q}(\thetav))+ f^{*}(-\frac{\foo{L}_{\dset{z}}(\thetav)+\beta}{\lambda}))(1 - \frac{\diff P_2}{\diff \Pgibbs{P}{Q}}(\thetav)) \diff Q(\thetav)\\
& = & \lambda\int ( f(\frac{\diff \Pgibbs{P}{Q}}{\diff Q}(\thetav))+ f^{*}(-\frac{\foo{L}_{\dset{z}}(\thetav)+\beta}{\lambda}))(\frac{\diff P_2}{\diff \Pgibbs{P}{Q}}(\thetav) - \frac{\diff P_1}{\diff \Pgibbs{P}{Q}}(\thetav)) \diff Q(\thetav).\label{EqGenAnyP_s3}
\end{IEEEeqnarray}
Substituting the probability measures $P_1$ and $P_2$ for the probability measures $P_{\Thetam|\bm{Z}=\dset{z}}$ and $P_{\Thetam}$; and taking the expectation of~\eqref{EqGenAnyP_s3} with respect to $P_{\dset{Z}}$ yields
\begin{IEEEeqnarray}{rCl}
\IEEEeqnarraymulticol{3}{l}{
   \overline{\overline{\foo{G}}}(P_{\Thetam|\bm{Z}},P_{\bm{Z}}) 
  }\nonumber \\
  & = & \lambda \int \int ( f(\frac{\diff \Pgibbs{P}{Q}}{\diff Q}(\thetav))+ f^{*}(-\frac{\foo{L}_{\dset{z}}(\thetav)+\beta}{\lambda}))(\frac{\diff P_{\Thetam|\bm{Z}=\dset{z}}}{\diff \Pgibbs{P}{Q}}(\thetav) - \frac{\diff P_{\Thetam}}{\diff \Pgibbs{P}{Q}}(\thetav)) \diff Q(\thetav) \diff P_{\dset{Z}}(\dset{z}),
\end{IEEEeqnarray}
which completes the proof.
\end{IEEEproof}

\subsection{Proof Theorem~\ref{theo_ERM_fDR_gen}}
\begin{IEEEproof}
\label{app_theo_ERM_fDR_gen}
From Theorem~\ref{Theo_f_ERMRadNik} and~\cite[Corollary 23.5.1]{rockafellar1970conjugate}, the Legendre-Fenchel transform in Definition~\ref{DefLT_cnvxcnj} satisfies for all $\thetav \in \supp Q$,
\begin{IEEEeqnarray}{rCl}
\label{EqLF_transf3}
     \dot{f}(\frac{\diff \Pgibbs{P}{Q}}{\diff Q}(\thetav))\frac{\diff \Pgibbs{P}{Q}}{\diff Q}(\thetav)& = &f(\frac{\diff \Pgibbs{P}{Q}}{\diff Q}(\thetav))+f^{\star}(-\frac{\foo{L}_{\dset{z}}(\thetav)+\beta}{\lambda}).
\end{IEEEeqnarray}
Then, from Theorem~\ref{Theo_ERM_fDR_LT_GenERr} and~\eqref{EqLF_transf3}, the generalization error of the solution to the ERM-$f$DR problem in~\eqref{EqOp_f_ERMRERNormal} satisfies
\begin{IEEEeqnarray}{rCl}
\IEEEeqnarraymulticol{3}{l}{
   \overline{\overline{\foo{G}}}(P^{(Q,\lambda)}_{\Thetam|\Zm},P_{\bm{Z}}) 
  }\nonumber \\
  & = & \lambda \int \int ( f(\frac{\diff \Pgibbs{P}{Q}}{\diff Q}(\thetav))+ f^{*}(-\frac{\foo{L}_{\dset{z}}(\thetav)+\beta}{\lambda}))(\frac{\diff \Pgibbs{P}{Q}}{\diff \Pgibbs{P}{Q}}(\thetav) - \frac{\diff \PgibbsNonCond{P}{Q}}{\diff \Pgibbs{P}{Q}}(\thetav)) \diff Q(\thetav) \diff P_{\dset{Z}}(\dset{z})\label{EqGE_GibbsP_s1}\\
  & = & \lambda \int \int ( f(\frac{\diff \Pgibbs{P}{Q}}{\diff Q}(\thetav))+ f^{*}(-\frac{\foo{L}_{\dset{z}}(\thetav)+\beta}{\lambda}))( 1 - \frac{\diff \PgibbsNonCond{P}{Q}}{\diff \Pgibbs{P}{Q}}(\thetav)) \diff Q(\thetav) \diff P_{\dset{Z}}(\dset{z})\label{EqGE_GibbsP_s2}\\
  & = & \lambda \int \int \dot{f}(\frac{\diff \Pgibbs{P}{Q}}{\diff Q}(\thetav))\frac{\diff \Pgibbs{P}{Q}}{\diff Q}(\thetav)( 1 - \frac{\diff \PgibbsNonCond{P}{Q}}{\diff \Pgibbs{P}{Q}}(\thetav)) \diff Q(\thetav) \diff P_{\dset{Z}}(\dset{z})\label{EqGE_GibbsP_s3}\\
  & = & \lambda \Bigg(\int \int \dot{f}(\frac{\diff \Pgibbs{P}{Q}}{\diff Q}(\thetav))\frac{\diff \Pgibbs{P}{Q}}{\diff Q}(\thetav)\diff Q(\thetav) \diff P_{\dset{Z}}(\dset{z}) 
  \nonumber \\ &  & - \int \int \dot{f}(\frac{\diff \Pgibbs{P}{Q}}{\diff Q}(\thetav))\frac{\diff \PgibbsNonCond{P}{Q}}{\diff Q}(\thetav) \diff Q(\thetav) \diff P_{\dset{Z}}(\dset{z})\Bigg)\label{EqGE_GibbsP_s4}\\
  & = & \lambda \Bigg(\int \int \dot{f}(\frac{\diff \Pgibbs{P}{Q}}{\diff Q}(\thetav))\diff \Pgibbs{P}{Q}(\thetav) \diff P_{\dset{Z}}(\dset{z})
  - \int \int \dot{f}(\frac{\diff \Pgibbs{P}{Q}}{\diff Q}(\thetav)) \diff \PgibbsNonCond{P}{Q}(\thetav) \diff P_{\dset{Z}}(\dset{z})\Bigg),\label{EqGE_GibbsP_s5} 
\end{IEEEeqnarray}
where~\eqref{EqGE_GibbsP_s3} follows from~\eqref{EqLF_transf3},~\eqref{EqGE_GibbsP_s4} follows from Corollary~\ref{coro_mutuallyAbsCont}.
This completes the proof.
\end{IEEEproof}

\subsection{Proof Remark~1}
\begin{IEEEproof}
Under the assumption that the function $f$ in~\eqref{EqOp_f_ERMRERNormal} is
\begin{IEEEeqnarray}{rCl}
\label{EqremarkAssum}
    f(x) & = &x \log(x),
\end{IEEEeqnarray}
from the Legendre-Fenchel transform in Definition~\ref{DefLT_cnvxcnj} it follows that
\begin{IEEEeqnarray}{rCl}
\label{EqremarkLFT4KL}
f^{\star}(t) & = & \exp(t+1).
\end{IEEEeqnarray}
Note that for the relative entropy, it also holds that
\begin{IEEEeqnarray}{rCl}
\frac{\diff}{\diff t}f^{\star}(t) & = & \exp(t+1),
\end{IEEEeqnarray}
which together with~\eqref{EqremarkLFT4KL} and Theorem~\ref{Theo_f_ERMRadNik} yields
\begin{IEEEeqnarray}{rCl}
\label{EqremarkLFT4KL2}
f^{\star}(-\frac{\foo{L}_{\dset{z}}(\thetav)+\beta}{\lambda}) & = & \frac{\diff \Pgibbs{P}{Q}}{\diff Q}(\thetav).
\end{IEEEeqnarray}
Then, under the assumption in~\eqref{EqremarkAssum}, the Gibbs algorithm satisfies for all $\dset{z}\in \supp P_{\dset{Z}}$ and for all $\thetav \in \supp Q$,
\begin{IEEEeqnarray}{rCl}
\label{EqremarRadNik}
\frac{\diff P_{\Thetam|\dset{Z}=\dset{z}}}{\diff Q}(\thetav) & = & \frac{\diff \Pgibbs{P}{Q}}{\diff Q}(\thetav).
\end{IEEEeqnarray}
Then, from Theorem~\ref{Theo_ERM_fDR_LT_GenERr} it follows that,
\begin{IEEEeqnarray}{rCl}
\IEEEeqnarraymulticol{3}{l}{
 \overline{\overline{\foo{G}}}(P^{(Q,\lambda)}_{\Thetam|\Zm},P_{\bm{Z}})
 } \nonumber \\
 & = & \lambda \int \int ( f(\frac{\diff \Pgibbs{P}{Q}}{\diff Q}(\thetav))+ f^{*}(-\frac{\foo{L}_{\dset{z}}(\thetav)+\beta}{\lambda}))(\frac{\diff \Pgibbs{P}{Q}}{\diff \Pgibbs{P}{Q}}(\thetav) - \frac{\diff \PgibbsNonCond{P}{Q}}{\diff \Pgibbs{P}{Q}}(\thetav)) \diff Q(\thetav) \diff P_{\dset{Z}}(\dset{z}) \label{EqremarkSimply_s1}\\
 & = &  \lambda \int \int ( \frac{\diff \Pgibbs{P}{Q}}{\diff Q}(\thetav)\log(\frac{\diff \Pgibbs{P}{Q}}{\diff Q}(\thetav))+ \frac{\diff \Pgibbs{P}{Q}}{\diff Q}(\thetav))(1 - \frac{\diff \PgibbsNonCond{P}{Q}}{\diff \Pgibbs{P}{Q}}(\thetav)) \diff Q(\thetav) \diff P_{\dset{Z}}(\dset{z})\label{EqremarkSimply_s2}\\
  & = &  \lambda \int \int (\log(\frac{\diff \Pgibbs{P}{Q}}{\diff Q}(\thetav))+ 1) \frac{\diff \Pgibbs{P}{Q}}{\diff Q}(\thetav)(1 - \frac{\diff \PgibbsNonCond{P}{Q}}{\diff \Pgibbs{P}{Q}}(\thetav)) \diff Q(\thetav) \diff P_{\dset{Z}}(\dset{z})\label{EqremarkSimply_s3}\\
   & = &  \lambda \int \Bigg(\int (\log(\frac{\diff \Pgibbs{P}{Q}}{\diff Q}(\thetav))+ 1)\frac{\diff \Pgibbs{P}{Q}}{\diff Q}(\thetav)\diff Q(\thetav)
   \nonumber \\ & &
   -\int (\log(\frac{\diff \Pgibbs{P}{Q}}{\diff Q}(\thetav))+ 1)\frac{\diff \PgibbsNonCond{P}{Q}}{\diff Q}(\thetav) \diff Q(\thetav)\Bigg) \diff P_{\dset{Z}}(\dset{z})\label{EqremarkSimply_s4}\\
    & = &  \lambda \int \Bigg[\int \log(\frac{\diff \Pgibbs{P}{Q}}{\diff Q}(\thetav))\diff \Pgibbs{P}{Q}(\thetav)
   -\int \log(\frac{\diff \Pgibbs{P}{Q}}{\diff Q}(\thetav)) \diff \PgibbsNonCond{P}{Q}(\thetav)\Bigg] \diff P_{\dset{Z}}(\dset{z}),\label{EqremarkSimply_s5}
\end{IEEEeqnarray}
where~\eqref{EqremarkSimply_s1} follows from~\eqref{EqremarkAssum} and~\eqref{EqremarRadNik}. Note also that~\eqref{EqremarkSimply_s4} is the result of substituting $\dot{f}(x)=\log(x)+1$ into Theorem~\ref{theo_ERM_fDR_gen}. This completes the proof.
\end{IEEEproof}

\end{document}